THE PENNSYLVANIA STATE UNIVERSITY
SCHREYER HONORS COLLEGE

DEPARTMENT OF ELECTRICAL ENGINEERING AND COMPUTER SCIENCE

FairLay-ML: Intuitive Remedies for Unfairness in Data-Driven Social-Critical Algorithms

Normen Yu
Spring 2023

A thesis
submitted in partial fulfillment
of the requirements
for baccalaureate degrees
in Computer Science
with honors in Computer Science

Reviewed and approved* by the following:

Gang Tan
Professor of Computer Science and Engineering
Thesis Supervisor

Danfeng Zhang
Associate Professor of Computer Science and Engineering
Honors Adviser

*Signatures are on file in the Schreyer Honors College.

Special thanks to:

Saeid Tizpaz-Niari
Assistant Professor of Computer Science
University of Texas at El Paso
Thesis Mentor

# Abstract

This thesis explores open-sourced machine learning (ML) model explanation tools to understand whether these tools can allow a layman to visualize, understand, and suggest intuitive remedies to unfairness in ML-based decision-support systems. Machine learning models trained on datasets biased against minority groups are increasingly used to guide life-altering social decisions, prompting the urgent need to study their logic for unfairness. Due to this problem's impact on vast populations of the general public, it is critical for the layperson – not just subject matter experts in social justice or machine learning experts – to understand the nature of unfairness within these algorithms and the potential trade-offs. Existing research on fairness in machine learning focuses mostly on the mathematical definitions and tools to understand and remedy unfair models, with some directly citing user-interactive tools as necessary for future work. This thesis presents FairLay-ML, a proof-of-concept GUI integrating some of the most promising tools to provide intuitive explanations for unfair logic in ML models by integrating existing research tools (e.g. Local Interpretable Model-Agnostic Explanations) with existing ML-focused GUI (e.g. Python Streamlit). We test FairLay-ML using models of various accuracy and fairness generated by an unfairness detector tool, Parfait-ML, and validate our results using Themis. Our study finds that the technology stack used for FairLay-ML makes it easy to install and provides real-time black-box explanations of pre-trained models to users. Furthermore, the explanations provided translate to actionable remedies.

Out of the twenty-four unfair models studied, we are able to provide a very clear explanation to four. Of the four, three lead to a clear increase in fairness for age, gender, and race across the models without decrease in accuracy. For example, FairLay-ML indicates that native country is used as a proxy to determine someone's race in one of the unfair models. In this example, FairLay-ML indicates that someone from South Africa is very likely not to be Caucasian, and the model decreases its prediction probability by 0.02 for someone from South Africa. We show that masking native country leads to a fairer model.

# Table of Contents

# List of Figures

# List of Tables

# Acknowledgements

I would like to extend my sincere gratitude to Dr. Gang Tan and Dr. Saeid Tipaz-Niari for their inspiration on this topic. Their consistent guidance throughout my undergraduate career is heartfelt. I am indebted to their patience and advice as I navigated between research, work, and school in the past few years.

I would also like to thank the Systems and Internet Infrastructure Security (SIIS) lab for providing a high-quality cloud-based virtual environment to develop, test, and run my code-base.

# Chapter 1

# Overview

## 1.1   Motivation

### 1.1.1   Problem Statement

Addressing systemic bias and unfairness in society is fundamentally a human problem, and nothing is more "systemic" in the 21st century than life-altering decisions made by automated algorithms. Oftentimes, these automated decision-making algorithms act as gatekeepers for many social decisions: should a person go to jail [1], is a person qualified for a loan [2], and more. This is especially concerning when many of these algorithms are becoming less human-driven and human-understandable, migrating to machine learning algorithms that learn the logic on their own. This problem of human understandability and explainability of machine learning algorithms will only exacerbate as we teach them to take more variables into account and their inner logic becomes more and more complex. To solve this problem, fairness in machine learning has become a hot topic, attracting the attention of many data scientists and machine learning experts.

For such a fundamentally human problem, the only interface between the world of society and the world of "fairness in machine learning" seems to be in the mathematical definition of fairness: true positive rates, average odd differences, data independence, data separation, data sufficiency, etc. Everything past that becomes merely a technical problem. To get a sense of the severity of the situation, one needs to look no further than the Wikipedia page for the field of fairness in machine learning[1]. The Wikipedia page focuses primarily on the definitions of fairness as they relate to ethical concerns and mathematical fairness mitigation tools based on the definitions. This point-of-view is echoed across the field. Indeed, one meta-study on the papers published in some of the most prominent conferences of AI ethics[2] explicitly concluded that there is "a great tendency for abstract discussion of such topics devoid of structural and social factors and specific and potential harms" [3]. This gap between bias in statistical terms and bias in political terms is not just an abstract difference between STEM and humanities. As the speed of technological growth increases exponentially, the lag between new artificial intelligence models, hypothetical methods to explain these models, and human integration to visualize the explanation will only increase. For such a problem with vast impacts on laypeople, it is critical for the general public – not just subject matter experts in social justice or machine learning experts – to understand the nature of unfairness within these algorithms and the potential trade-offs for potential remedies.

Unfortunately, the current open-sourced technologies are not positioned for human-machine-integration, with some publications directly encouraging future studies to integrate their proposed tools for human interactivity. Highly integrated tools – such as Amazon SageMaker and Snorkel AI – are often proprietary and too costly for the general population (see Section 1.1.2 below). *This thesis explores data visualization tools and open-sourced machine learning (ML) model expla-nation tools – including Streamlit, Plotly, and Local Interpretable Model-Agnostic Explanations – to understand whether they can allow a layman to visualize, understand, and suggest intuitive remedies for unfairness in machine learning models.*

---

[1]The Wikipedia page can be found in `https://en.wikipedia.org/wiki/Fairness_(machine_learning)`

[2]The study, published in 2022, focuses on the papers published in the AIES and FAccT conferences.

### 1.1.2 Existing Technology

From a human-centric angle, Grafana is leading the pack in rapid, drag-and-drop style data visualization. However, Grafana works on servers, integrating with SQL databases and web servers for visualization. Consequently, setting up configurations and interfaces for Grafana requires significant upfront costs. Furthermore, Grafana's configuration is not built for portability, making it difficult to share visualization tools among computers and collaborators.

Tools such as Amazon SageMaker and Snorkel AI have also emerged to bridge the gap between visualizing the data in training, analyzing the model, A/B testing, and performance metrics (including bias, unfairness, and accuracy) in deployment. These MLOps tools bridge the deployment gap for machine learning algorithms development into IT operations. Unfortunately, Amazon SageMaker is proprietary and requires the continuous purchase of Amazon's computing power, services, and infrastructure to use. Meanwhile, Snorkel AI is an amazing tool to enforce less bias in datasets by interfering in the labeling process itself. To mitigate bias, Snorkel AI allows analysts and subject matter experts to inject their expertise into the data pipeline to better train machine learning algorithms. Unfortunately, Snorkel AI assumes a subject matter expert whom we can ask for justification. These constraints make it infeasible for a regular, concerned citizen (e.g. a regular voter in an election) to use and gain a better understanding of fairness in AI.

A variety of tools in the open-source community are being developed to address some of these issues. From a mathematical angle, a Python toolkit called FairLearn is leading the path towards a completely open-sourced, standard library to analyze data and machine learning models on the most popular definitions of fairness. This toolkit is modularly separate and compatible with most machine-learning Python libraries such as scikit-learn, PyTorch, TensorFlow, and more. These tools integrate through Python data objects (e.g. Numpy and Pandas). A Python-based human-centric tool, Streamlit, has also emerged to support web-based visualization that is easily programmable with a few lines of Python code. Of course, this code can be shared through standard ways such as GitHub, allowing Streamlit servers to be shareable easily without more advanced infrastructures such as Docker.

On the other side of the academia-industrial spectrum, many papers have proposed the use of visualization or model-agnostic explainability technologies, some on similar technologies that we will propose in Section 2 [4, 5]. However, these papers do not position the technologies as software packages for integration. Indeed, one such paper directly notes this lack of integration, stating in its future studies " it would be really useful to embed our proposed methodology in user-interaction tools and perform studies both to validate our method and also to improve it by taking into account user feedback, possibly allowing the users to change, among other parameters, the feasibility constraints on actions" [5].

## 1.2 Background

We build our visualization toolkit FairLay-ML based on Parfait-ML – a search-based software testing software that can be used to generate fairer machine learning models – with the understanding that the concepts tested should be extendable to any machine learning models and datasets [6]. Parfait-ML provides a concrete baseline to allow us to analyze our visualization tool, as it generates *classical models* – logistic regression, decision tree, support vector machine, and random forest – that are easy to visualize. As illustrated in Figure 1.1, FairLay-ML represents the key logic of the decision tree and random forest models with tree graphs, and the key logic of logistic regression and support vector machine with simple bar plots. Users can intuitively imagine the tree graph as conditional statements for feature values, and the bar plots as a weighted sum of the features values. Of course, some details such as the cutoff thresholds or the bias terms are abstracted away. Still, the overall direct explanation for how a model classifies data points is clear.

Figure 1.1: Visualization of Classical Models

(a) Visualization of Decision Tree Models

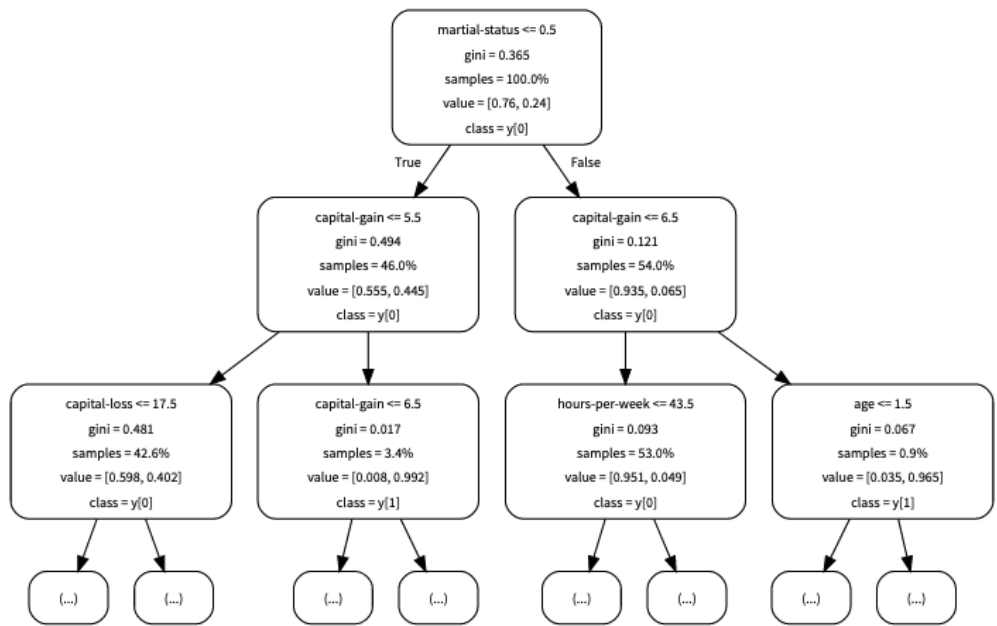

The logical representation of a decision tree is provided as a tree graph. Each data point starts from the top node and traverses through the sub-branches based on the values of each feature for the data point. In this way, each internal node of the tree acts as a conditional expression to determine the sub-branch until a leaf node is visited, which specifies the classification. Due to the high number of nodes that contribute to the final classification, users choose the maximum depth to display. The confidence score of the prediction is the fraction of training data points of the same classification in the leaf node [7].

(b) Visualization of Random Forest Models

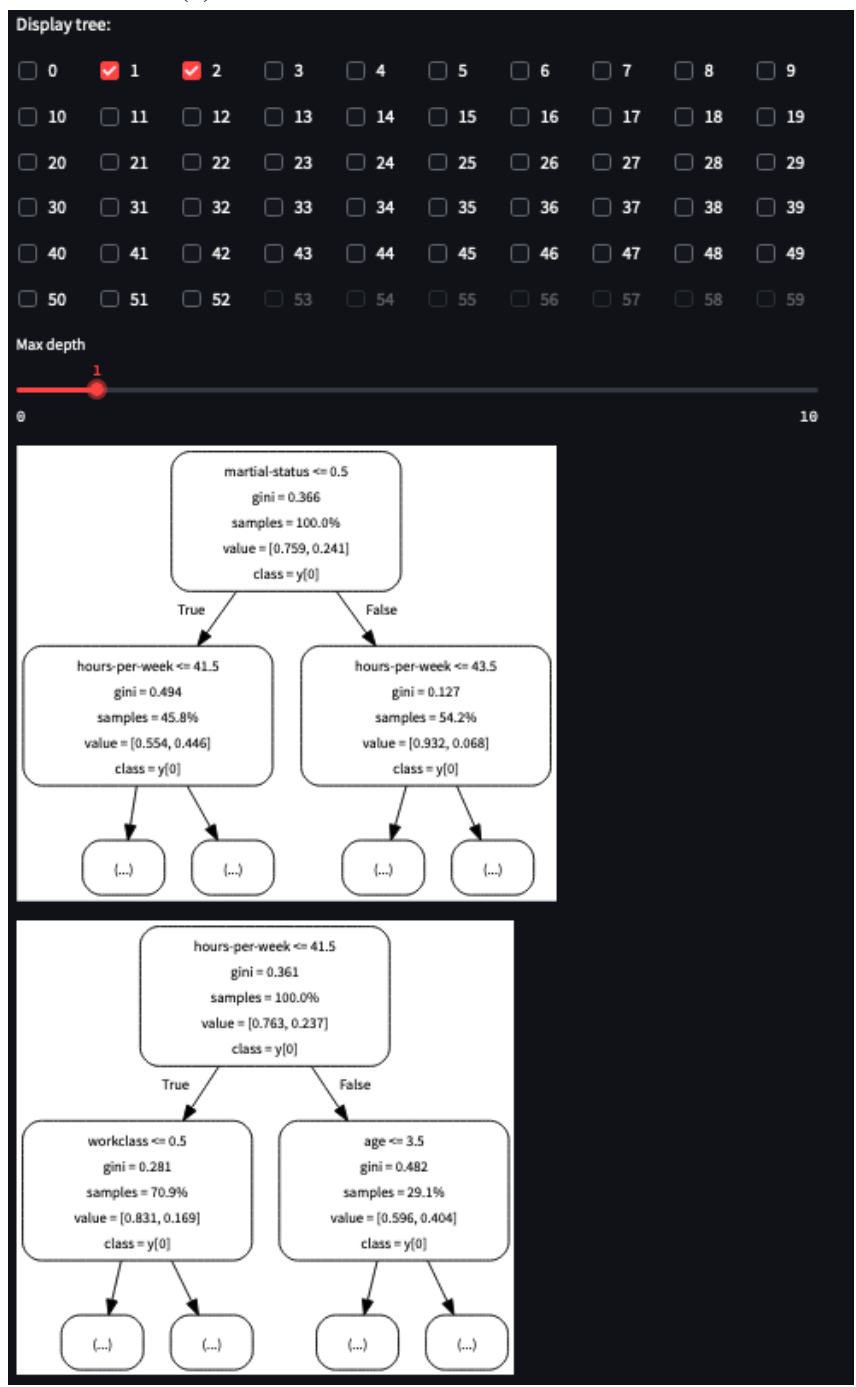

The logical representation of a random forest is based on decision trees: a data point is fed through a number of decision trees. Each decision tree provides a classification. The classification with the most "votes" from all of the trees determines the final classification. Due to the high number of tree logic that contributes to the final voting, users choose the trees to display. We display and represent each decision tree logic identical to a decision tree model (see Figure 1.1a above). The prediction is generated by voting, with equal weights for all trees, and the confidence score is the average of the confidence scores of each tree [7].

(c) Visualization of Support Vector Machine Models

**The SVM Model's standard deviation adjusted hyperplane slopes**

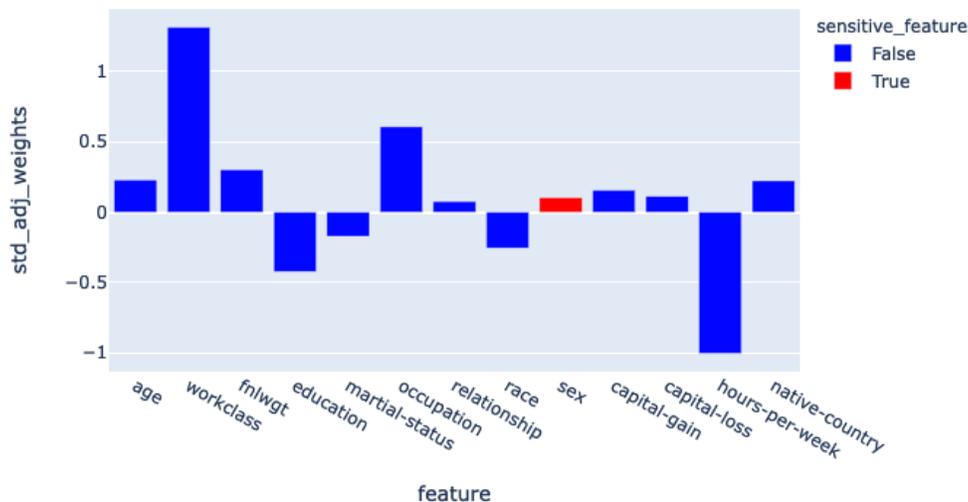

The logical representation of support vector machines is provided as the weights for each feature that represents the hyperplane. We provide the actual weights (not shown) and the standard deviation adjusted weights (shown) for each feature in two separate graphs. The classification of the model for a data point is determined by which side of the hyperplane the data point resides. The confidence score for the prediction is using cross-validation [7]

(d) Visualization of Logistic Regression Models

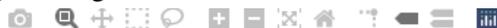

**The LR Model's standard deviation adjusted weights**

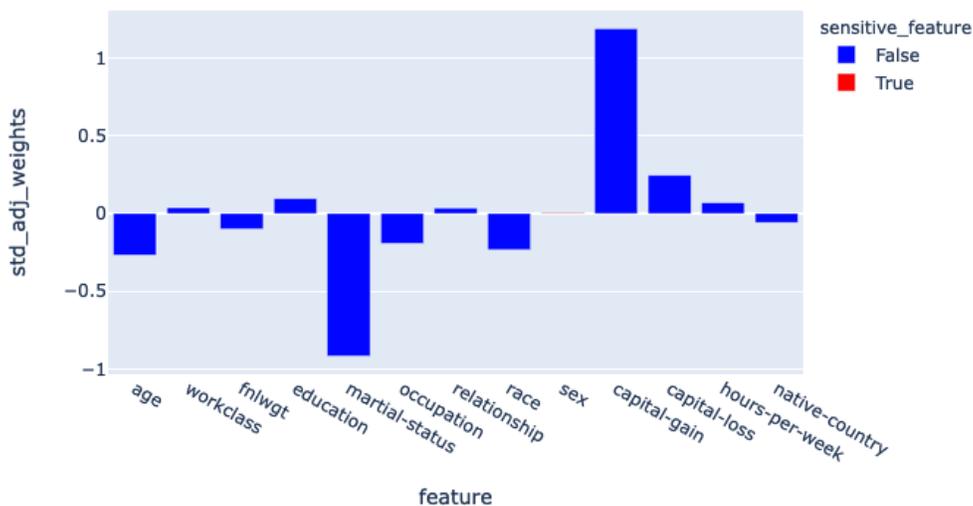

Logistic regression uses a weighted sum of a numerical encoding of the features. FairLay-ML shows the weights of the weighted sum, providing both the actual weights (not shown) and the standard deviation adjusted (shown) weights. The final numerical value provides a classification using a threshold (if value smaller than a threshold, class 0, else class 1) and a confidence score by normalizing the value [7].

### 1.2.1    The Datasets

Parfait-ML trains a variety of classical models using tabular bench-mark fairness datasets such as the German Credit Data, Adult Census Income Data, Bank Marketing Data, and COMPAS recidivism dataset [8, 9]. Each of these datasets contains around ten to twenty columns. Some feature columns represent numerical data types while other columns represent categorical data. Columns representing numerical data types were unchanged, while columns representing categorical data were numerically encoded such that scikit-learn models can treat them numerically. Each of the datasets contains more than 1,000 data points, with the Bank Marketing dataset and the German Credit dataset having more than 30,000 data points. Some of the feature columns included could be considered sensitive, such as someone's age, gender, or race. For simplicity, we consider the sensitive features to have only two categories[3]. Each dataset also contains a label column, which is ultimately the feature that we would like our models to predict. Labels are also binary categories numerically encoded as a 0 or 1 for each data point:

- for the Census dataset, they represent less than 50k annual income and more than 50k annual income, respectively;

- for the Credit dataset, they represent whether a client seeking a loan is not risky or risky;

- for the Bank dataset, they represent whether an existing client has not subscribed or already subscribed to a product;

- for the Compas dataset, they represent whether a suspect or inmate represents a low risk of recidivating or a high risk of recidivating.

### 1.2.2    Fairness Definitions

Due to inherent unfairness and correlations in the datasets [10], a model trained on the dataset typically carries bias and unfairness against certain sub-populations. Parfait-ML studies and mitigates this issue by randomly mutating model hyperparameters and studying the fairness of each model. Therefore, by nature, Parfait-ML generates a large number of learned models with varying inherent fairness.

In order to simplify the study, we study fairness in machine learning with respect to one feature at a time. The feature of concern, such as gender, race, or age, is often called the *sensitive feature*. In cases where there are multiple sensitive features within the dataset, we create multiple case-studies to investigate each sensitive feature independently. Parfait-ML calculates a fairness score using a statistical separation metric called *average odd difference*, which is based on four very simple metrics. For a categorical, sensitive feature D and a category s in the feature[45]:

- False Positive (FP) with respect to a feature[6] $D = s$ is the number of people in s that were categorized as "positive" but are actually not. For example in the context of the German credit ratings, False Positive of female is the total number of females whom our algorithm mistakenly classified as risky but are actually not risky.

---

[3]For age, we consider only people under 25 as the first category and over 45 as the second, disregarding anyone that does not fall into these two ranges. For race, we consider only Caucasians and everyone else as "non-Caucasians". For gender, we only consider males and females.

[4]A numerical sensitive feature can be binned to multiple categories.

[5]e.g. For example, a category for race is caucasian.

[6]A subscript of $D = s$ refers to the sub-population where all data points into category s for the feature D.

- True Positive (TP) with respect to a feature $D = s$ is the number of people in s that were correctly categorized as "positive". For example in the context of the German credit ratings, True Positive of female is the total number of females whom our algorithm classified as risky are actually risky.

- False Negative (FN) with respect to a feature $D = s$ is the number of people in s that were incorrectly categorized as "negative". For example in the context of the German credit ratings, False Negative of female is the total number of females whom our algorithm classified as not risky but are actually risky.

- True Negative (TN) with respect to a feature $D = s$ is the number of people in s that were correctly categorized as "negative". For example in the context of the German credit ratings, True Negative of female is the total number of females whom our algorithm correctly classified as not risky.

Based on these values, we can then calculate True Positive Rates with respect to a feature D=s and False Positive Rates with respect to $D = s$:

$$\text{TPR}|_{D=s} = \frac{\text{TP}|_{D=s}}{\text{TP}|_{D=s} + \text{FN}|_{D=s}} \tag{1.1}$$

$$\text{FPR}|_{D=s} = \frac{\text{FP}|_{D=s}}{\text{FP}|_{D=s} + \text{TN}|_{D=s}} \tag{1.2}$$

**Group Fairness**

Parfait-ML determines the fairness of each model using *average odd difference* (AOD), or the average of differences between the true positive rates (TPR) and the false positive rates (FPR) of two protected groups. The calculation for the average odd difference is shown in Equation 1.3[7][11].

$$\text{AOD}|_{D=\text{category\_1}} = \frac{(\text{FPR}|_{D=\text{category\_1}} - \text{FPR}|_{D=\text{category\_2}}) + (\text{TPR}|_{D=\text{category\_2}} - \text{TPR}|_{D=\text{category\_2}})}{2} \tag{1.3}$$

In Section 2.6, we independently verify the fairness results of Parfait-ML using a fairness testing tool called Themis [12]. Themis uses two different metrics to evaluate the fairness of models: group discrimination score and causal discrimination score.

The *group discrimination* score is calculated based simply on the difference in the fraction of classified positives among the difference sensitive groups. Equation 1.4 provides the equivalent equation for Themis definitions based on established metrics from Parfait-ML.

$$\text{group discrimination} = \frac{\text{FP}|_{D=\text{category\_1}} + \text{TP}|_{D=\text{category\_1}}}{\text{FP}|_{D=\text{category\_1}} + \text{TP}|_{D=\text{category\_1}} + \text{FN}|_{D=\text{category\_1}} + \text{TN}|_{D=\text{category\_1}}}$$
$$- \frac{\text{FP}|_{D=\text{category\_2}} + \text{TP}|_{D=\text{category\_2}}}{\text{FP}|_{D=\text{category\_2}} + \text{TP}|_{D=\text{category\_2}} + \text{FN}|_{D=\text{category\_2}} + \text{TN}|_{D=\text{category\_2}}} \tag{1.4}$$

[7]Recall for simplicity of the study, we only consider 2 sensitive categories for each case-study

This definition is founded on a communal approach, as the fairness of the overall groups of people is emphasized.

**Causal Fairness**

The other metrics used by Themis follow a more causal philosophy. The Themis' *causal discrimination* score is calculated based on *counterfactuals*. A *counterfactual* of a classification model points in the feature space where the model's classification output changes if the sensitive feature is changed. The set of counterfactuals of a classification model, given a test dataset, is the subset of the test dataset that are counterfactuals. The causal discrimination score of a model is simply the percentage of a test dataset that is in the set of counterfactuals. Parfait-ML is also capable of providing this metric by substituting using the training dataset.

This definition is founded on a more individualistic approach, as fairness to each individual person is emphasized.

# Chapter 2

# Methodology

## 2.1 Solution Statement

Our proof-of-concept solution, FairLay-ML, seeks to help humans make smarter, more informed decisions on how to best mitigate unfairness in machine learning models by leveraging and bridging the viewpoints and technologies specified in Section 1.1.2. To this end, the natural first step and scope of our solution are to increase human awareness of (1) data biases and (2) the hidden logic behind machine learning models. We achieve this complicated task[1] in an "intuitive" manner by following one simple principle across our solution: summarize results and provide more details interactively. Section 3.3.2 details how we utilize this design philosophy.

By integrating and leveraging some of the tools in the Background section (Section 1.2), we seek to offer an intuitive and clear picture of the datasets and machine learning models. Visualizing high-dimensional data is addressed interactively to provide a clearer understanding of how each feature dimension correlates with sensitive attributes (see Section 2.2 for details). This is used in tandem with the model visualization toolkit, which leverages Local Interpretable Model-Agnostic Explanation (LIME) toolkit to isolate the categories within features that impact the output the most (see Section 2.3 for details). This provides insights into how a machine learning algorithm could directly or indirectly discriminate against certain groups of individuals through correlations in the dataset. Tools are becoming more readily available to visualize the logic behind machine learning models. FairLay-ML pools these resources into a coherent, interactive infrastructure in order to support understanding and analysis of unfairness in machine learning. In Section 3, we dive into how these tools are implemented.

Ultimately, we use FairLay-ML to identify and remedy unfairness through three means: data features, model logic, and data sampling for model explainability. First, the data visualization tools provide an overview of how features are correlated to the sensitive feature. This tool helps users explore the distribution of classifications for each feature, as well as understand how the features in a dataset are interconnected in the context of bias in machine learning. FairLay-ML's data visualization tool identifies key features that we should pay attention to in our analysis of bias. For example, zip code has been found in many contexts to accidentally be used by ML algorithms as a proxy for race [13]. Second, FairLay-ML displays a model's logic (see Figure 1.1) to understand how a model uses the features to make decisions. Finally, FairLay-ML explains how a model classifies specific data points using LIME. LIME shows how each feature value of a data point impacts the model's decision.

These methods of visualization combine to provide explanations of the form "This model places a high weight on feature category X, and X is strongly correlated with the sensitive feature, causing unfairness;" this explanation is transformed into a remedy by masking category X in the feature and retraining the model (see Section 2.3 and Section 2.4). We evaluate the remedies by comparing accuracies/AOD scores and validate the results using Themis.

In summary, as illustrated in Figure 2.1, we first identify the most impactful categories within features that determine model prediction. We then check the dataset to see if these categories are indicative of sensitive attribute categories. If so, we mask the categories, retrain the same model, and observe any changes in accuracy or fairness scores in the training dataset and with a pseudo-testing dataset using Themis. Additional details for each step are provided below.

---

[1] We assert that this task is difficult because of the high dimensionality of the data and, by definition, machine learning models are used when direct and intuitive algorithms are not found.

Figure 2.1: Process Diagram

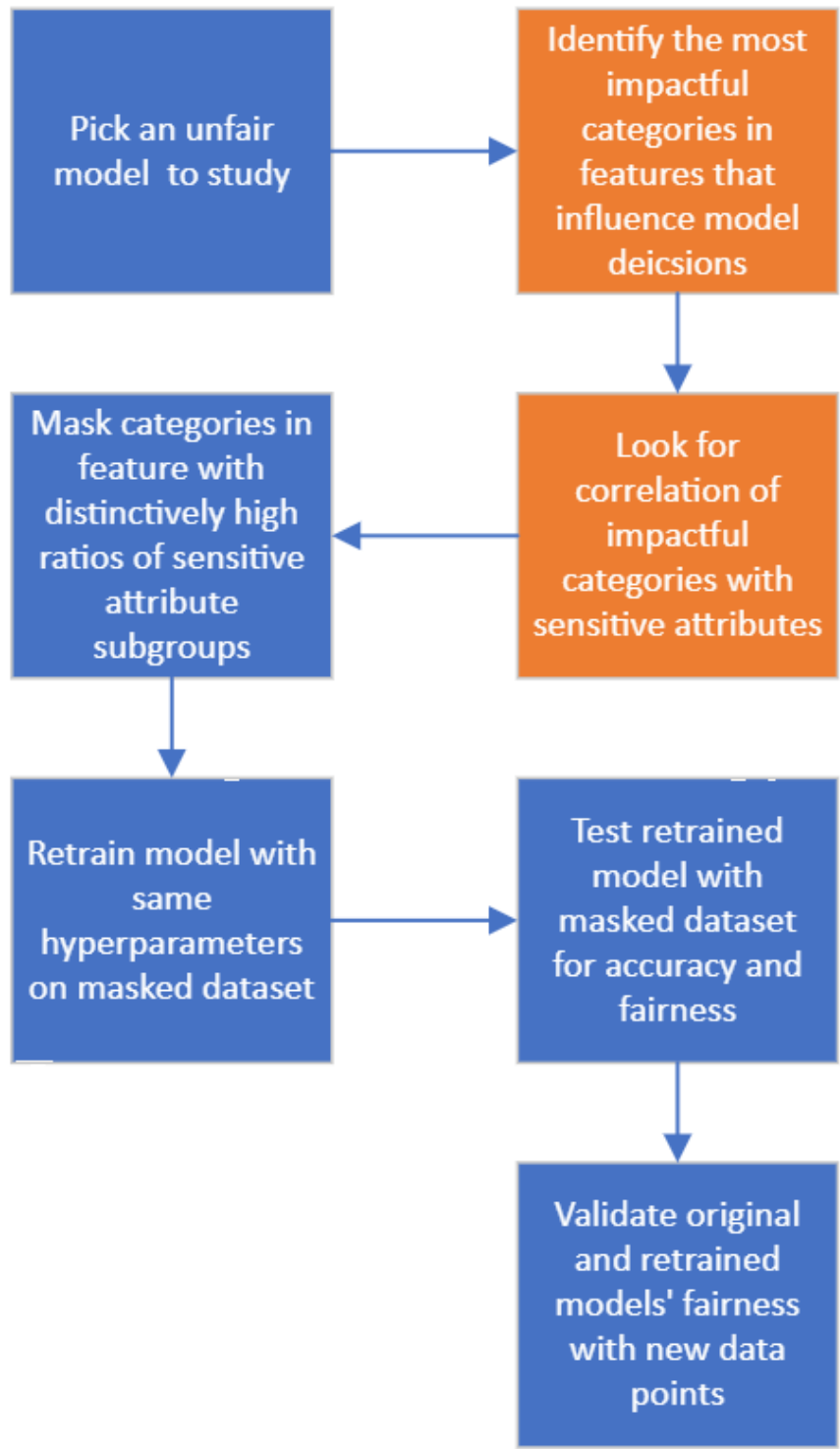

This diagram summarizes the key areas of work that led to the most impactful findings. The orange boxes indicate the process that is completed using the FairLay-ML GUI.

## 2.2 Direct Visualization of the Dataset and Models

Our first attempt to increase the explainability of trained models is to simply find ways to summarize the training data and visualize model logic directly. During this stage, we seek to increase our understanding of the datasets that are used to train the models and the models that are trained on the datasets. The fundamental problem with machine learning and artificial intelligence lies in the fact that the feature space (inputs) and the optimization space (trained weights) are of high dimension; this makes their direct visualization almost impossible. The natural first step – if possible – is therefore to find intuitive ways to visualize them on a 2D screen. Fortunately for the test cases of Parfait-ML, the feature space is only between 10-D to 20-D, and the optimization space is similar in size. For much higher dimensional applications such as images, language datasets, and deep neural networks, visualizing both the feature space and network with regard to fairness will present further difficulty that falls beyond the scope of this thesis.

We dive into more sophisticated methods to explain unfairness in Section 2.3. However as can be seen in Section 4.1, we refer to these more fundamental graphics to both validate our explanations and to provide intuition when our more sophisticated methods do not provide strong statistic responses.

### Visualizing the Dataset

In order to gain a better understanding of the dataset with respect to fairness, we create a custom data monitor in Streamlit that can analyze the correlation between each feature column with the sensitive feature. However, correlation turns out to be a bad metric for categorical features. While a strong correlation is a good indicator of possible unfairness, arbitrary numerical encoding methods mapping categorical features to arbitrary numerical values can mask the correlation metric. For a better analysis on the categories within a feature that is indicative of sensitive attributes, the exact distribution of categorical features condition on sensitive attributes is needed (see Section 3.4). This information proves very valuable during our discussion of the explainability of the models Section 4.1.

### Visualizing the Models

One of the first basic steps we take towards understanding the model is visualizing the model logic. We seek to gain a better understanding of models generated by Parfait-ML and connect these results with the feature correlations above (Section 2.2). The classical models studied by Parfait-ML – Logistic Regression, Decision Trees, Random Forests, and Support Vector Machine models – are pretty straightforward to represent compared to more modern deep neural networks: Section 3.5 provides exact details for how each of these models is visualized.

## 2.3 Explaining Unfairness in Models

Unfortunately, it turns out that simply representing the model logic is not sufficient in providing intuitive explanations and remedies for unfairness. Firstly, "slightly" unfair models can be caused by significant unfairness to a subset within a sensitive attribute[2]. In this way, only some categories within a feature are causing unfairness while others are less impacted. Secondly, decision tree and random forest models can use the same feature across many nodes, making the impact of the feature unclear.

Model explanation using LIME provides remedies to these problems. LIME takes in a model and a specific point in the feature space. It then perturbs the point and calculates the change in prediction probability for each feature. The changes in prediction probability for each feature are then plotted (see Section 3.6 for additional details). By randomly sampling a number of points, we can see how specific categories within features impact model prediction probabilities.

For our study of understanding model unfairness, we randomly sample 15 counterfactual points for each case-study. These points mostly have predictive probabilities near 0.5. Therefore, their final classification labels are more likely impacted by probability changes compared to a data point that the model very confidently (near 0 or 1 predictive probability) predicts to be a certain class.

## 2.4 Providing Remedies

By leveraging FairLay-ML, we achieve intuitive explanations for unfairness in models. From model explanations, we find categories in features that impact prediction probabilities the most. From data visualizations, we find categories in features that are strongly correlated to a data point's sensitive attribute. Combining these two results, our explanation comes in the form "This model places a high weight on feature category X, and X is strongly correlated with the sensitive feature, causing unfairness," where X is a category within a feature (e.g. "South Africa" for native country, or "wife" for relationship status).

Given such an explanation, our remedy is to mask the categories within a feature that are strongly correlated to the sensitive feature that also influences model decisions. Much past literature has cited that fairness-through-unawareness does not work precisely because of these proxies. [14, 15]. As indicated in Section 3.2.3, we also seek to confirm this work. From this perspective, we attempt to mask the proxies directly. We do this by collapsing all of the categories we seek to mask to the same value. We are selective in the categories that we mask to avoid masking too much valuable information from the model. We keep the feature column even if the whole feature is masked, as we seek to test out the model with the same architecture. We retrain the model using the same architecture and hyperparameters using the masked dataset and test our updated model using a masked dataset as well for fairness metrics.

## 2.5 Testing Remedies

We utilize models generated by Parfait-ML to study our infrastructure's ability to provide explainability for why models are fair or unfair and propose possible remedies. We also use Parfait-ML's calculated average odd difference metric to pick out good models to test: models that are most unfair, models that are most fair, and models that are most accurate. In order to demonstrate

---

FairLay-ML's explanation and remedies, we require unfair models. Therefore, we pick the most unfair models from Parfait-ML to study.

For each remedy, we observe the changes in accuracy and average odd difference score change for the picked model.

## 2.6   Validation with Themis

Up to this point, our study of explainability and remedies are based on the same training dataset. In general, this practice weakens the confidence of the results generated, as this provides opportunities for model overfitting and accidental cherry-picking. We utilize Themis [12, 16] to test the Parfait-ML models and remedied models generated. The Themis test provides a group discrimination score and a causal discrimination score, which we can use to compare the original model and the rememdied model. This alleviates concerns for multiple reuses of the training dataset since Themis generates its own random samples from the feature space to test the model.

Themis' randomly sampled points do not come with labels, necessitating definitions of fairness which can be computed without them. The two alternative definitions of fairness used by Themis, group score and causal score, span two different philosophical schools of thought regarding fairness. The exact philosophical and mathematical differences are highlighted in Section 1.2.

# Chapter 3

# Implementation

## 3.1    Implementation Summary Statement

The workflow starts with forking the GitHub repository from Parfait-ML [6] and making the necessary updates to save the models and the metadata necessary for future visualization and explanation study. We then build out FairLay-ML to gain a better understanding of the biases and correlations in the data, followed by tools to directly view the logic of the classical models that we seek to study. We integrate LIME to visualize explanations of the models and study the most fair, most accurate, and most unfair models for each study. Based on the visualizations, explanations and remedies are proposed to account for the unfairness. We then implement the remedies, retrain the same model, and compare fairness metrics to test our explanations and remedies. To avoid the reuse of the training dataset, we integrate the Themis fairness test to verify our results. The most impactful stages of our process, whose results will be discussed in Section 4, are summarized and highlighted in sequential order in Figure 2.1.

All of these tools are available as Python pip packages. Hence, any data engineer, data scientist, or machine learning engineer familiar with Python can simply pip installs each of these packages and run the tool as a web service. The data pipeline is integrated seamlessly through the commonly used Python Numpy and Pandas library, which comes with a very powerful, well-documented, memory and time-efficient library to manipulate any data set. The visualization and user interface pipeline, including Plotly and LIME, is integrated through Streamlit and displayed through a web server.

A high-level view of how each component should work with its surrounding infrastructure and the user is summarized in Figure 3.1. Our server is hosted by `https://streamlit.io/cloud`, which pulls the code, data sets (our Census, Bank, Compas, and Credit data sets), and pre-trained models (from Parfait-ML) from GitHub. Each model comes with meta-data that stores the accuracy, AOD, and hyperparameters of the model. All other graphs and metrics are calculated on the spot. Our infrastructure is split into the data visualization tool and the model visualizer. When a user enters the web app through `https://fairlay-ml.streamlit.app/`, they navigate to one of the two tools. For the data visualizer, the user first picks a data set. Then, a bias summary of the data set is provided, with additional details that display interactively as detailed in Section 3.4 below. For the model visualizer, the user first picks a case study. A Pareto-frontier scatterplot for that case-study of models with different hyperparameters are displayed, organized by the AOD and accuracy scores. The user can choose a specific model to view, which will prompt the server to show additional model logic (see Section 3.5), how the model predicted each training data, and additional explanations (see Section 3.6 for details).

The entire code base can also be pulled from GitHub to be run on a native computer (with graphics provided on a browser through localhost). The installation process and running of the code can be completed in less than 10 lines of shell commands.

Figure 3.1: FairLay-ML System Block Diagram

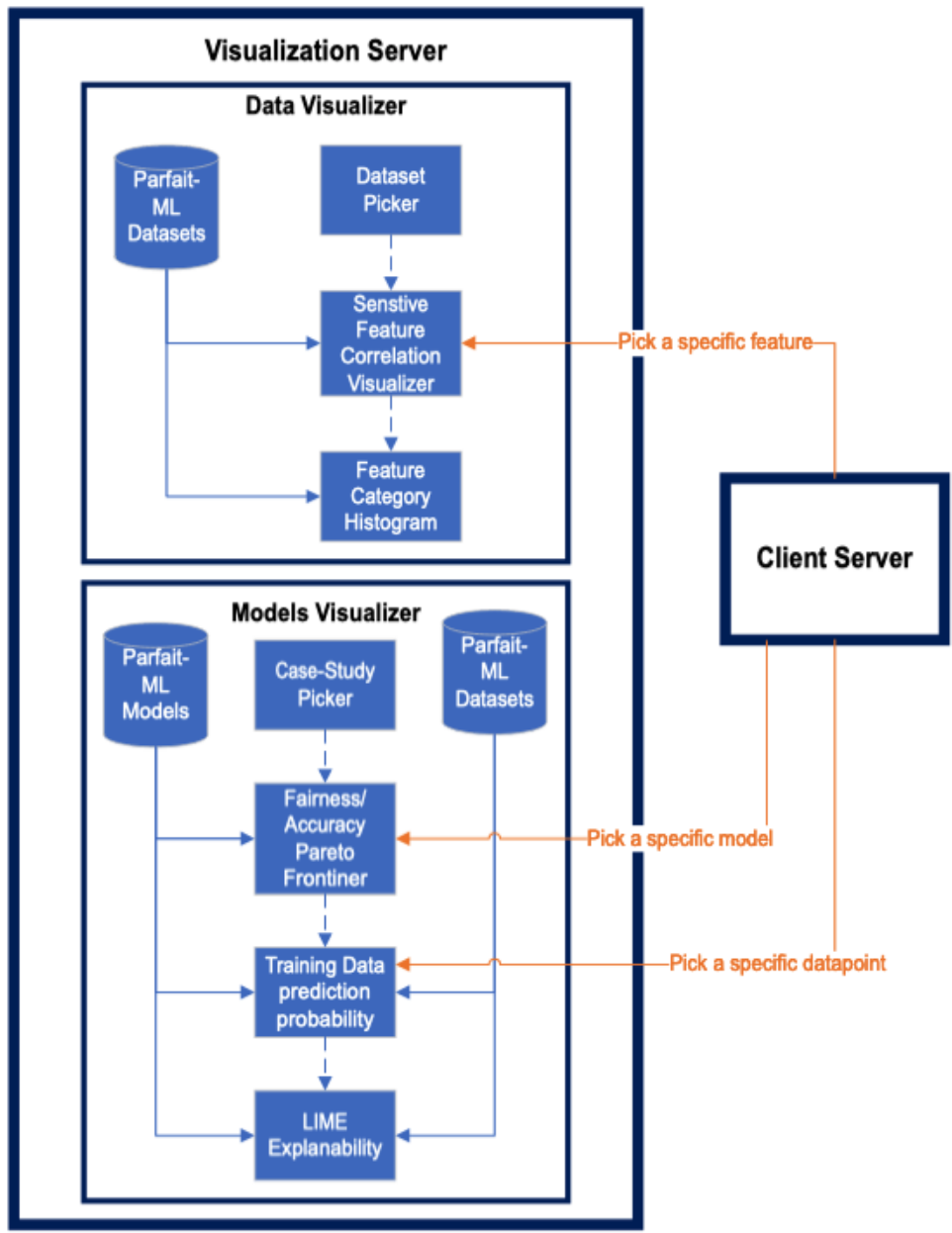

Our visualization tool is split into two sections: a data visualizer to investigate bias in data, and a model visualizer to investigate bias within the logic of Parfait-ML's trained models. The data and models are stored in a database. All other graphs and metrics are computed by the server in real-time, as requested by clients.

## 3.2    Integration with Parfait-ML

In Section 1.2, we discussed Parfait-ML's ability to create a large number of models across varying datasets, varying model types, and varying inherent fairness scores using randomly mutated hyperparameters. On top, Parfait-ML's ability to label each model's fairness and accuracy makes these models the ideal test cases for our proof-of-concept.

We updated a few components of Parfait-ML to allow for FairLay-ML to analyze and explain each model. After careful testing and scrutiny, the updates that we deemed most beneficial to the Parfait-ML open-source community were sent as a pull request. Careful compatibility and risk factors were considered during the opening of pull requests of each update to maximize the positive impact of the open-source repository while ensuring that the code still works the same way as described in the original publication. To be specific, the default configuration of any update must be backward compatible such that the results on the originally accepted paper on Parfait-ML [6] must be reproducible in the code's default configuration.

### 3.2.1    Saving Each Learned Model

In order to allow for models to be analyzed in FairLay-ML, especially to display the logic and predict new data points, each model must be saved as a Python pickle file on disk and accessible to the backend of our web server. Each saved model is titled with the hyperparameters that were used to create the model, and the file content contains the logic of the model (see Figure 3.2). An optional flag – defaulted to not saving the model – was added to trigger this update in order to preserve backward compatibility.

Figure 3.2: Learned Models

Each learned model generated by Parfait-ML is saved, with hyperparameters embedded in the title of the pickled file.

### 3.2.2    Dataset Labeling

Since machine learning algorithms often require feature vectors to be encoded into numeric categories, the datasets provided did not come with labels. The original source for the German Credit Data, Adult Census Income Data, and Bank Marketing Data all came with labels [8]; however, we were unable to confidently re-label all the feature columns of the dataset of COMPAS [9]. We were able to encode meaningful feature column labels directly into the dataset file while maintaining backward compatibility and minimal code changes due to scikit-learn's adaptability.

As discussed in Section 1.2, some feature columns fundamentally represent numerical data while others are simply the numerical encoding of categorical data. In order to facilitate human interpretability and explainability of any models, understanding and visualizing the numerical encodings is essential. Unfortunately, scikit-learn requires the data to be in numerical form. Thus, changing the numerical representation of the categorical feature columns causes significant downstream effects. In order to mitigate this risk, we created an additional Python module that distinguished categorical data columns from numerical data columns and numerical mappings to useful representation – e.g. 1 represents "married", while 0 represents "divorced" in marital status columns – were created. This module is now available to the community but not directly integrated into the pipeline.

### 3.2.3   Other Miscellaneous Updates

During the exploratory stages of the research, options including standard-scaling of the data were tested. This made the weights of the logistic regression model more meaningful to explainability, as a small weight on a data column with high variance can have the same impact on the final classification as a large weight on a data column with a small variance. Fairness through unawareness – the masking of the sensitive attribute from model training and testing – was also implemented and tested. While these updates ultimately yielded inconclusive results, these updates were vigorously tested, a pull request was made, and the original developers of Parfait-ML decided to integrate these options into their code base.

### 3.2.4   Updates that Were Not Approved

Not all updates were pulled into the original repository of Parfait-ML. The most common reason is that some updates, while possibly increasing quality or replacing deprecated components, carry strong risks of altering results, making it harder for future researchers to reproduce and verify results of the Parfait-ML paper [6]. One such update is to change Support Vector Machine from scikit-learn's LinearSVC to the regular SVC, which has the ability to utilize the linear kernel as the LinearSVC does as a special case of its hyperparameters. Scikit-learn's regular SVC also differs from LinearSVC in its ability to output a confidence score on its prediction; this is a necessary metric for some of the explainability tools we aim to test.

## 3.3   FairLay-ML Design Considerations

The design philosophy of FairLay-ML's infrastructure revolves around flexibility and usability. While the code is intended for a proof-of-concept using tabular bench-mark datasets as discussed in the Background section (Section 1.2), the technology tested should be intended for generalizability. The solution should be extendable to any dataset, any preprocessing tool, any machine learning model, and – most importantly – anyone with a computer. This leads to the following design constraints:

1. The server infrastructure should be easily shareable and usable

2. The code infrastructure should be compatible and tested on a variety of datasets and machine learning models

3. The outputs should be graphically informative for a layman (i.e. non machine learning or subject matter experts), intuitive to use, and interactive to show the desired information without overwhelming a user.

### 3.3.1 Decreasing Upfront Set-up Costs with Streamlit

These constraints lead to a very obvious candidate for visualization: Streamlit. Streamlit is a pip package that can be used to create web applications based solely on Python. Streamlit leverages the full flexibility, ease of use, and power of Python and the user interface of web browsers. There is no need for a complicated set-up of architectures such as SQL, nor memory-intensive installation of commercial-grade tools. Streamlit leverages the Python community's open-source packages equipped to handle any machine learning models and data types, compute common data manipulation techniques, and provide fast feedback to developers. As shown in Figure 3.3, a Streamlit Python script generates a web server; this leverages web browsers' GUI to provide an interactive display for users and minimizes additional installation of graphical packages. Developers looking to quickly and easily visualize their data and rapidly distribute their visualization tools can do so.

Of course, this comes at the sacrifice of run-time efficiencies. The lack of builds and compilations necessitates interpreted language, which runs slow by industrial standards. Furthermore, Streamlit kicks off a new Python process in the back end for every new user, which impedes scalability to a large number of users. Still, FairLay-ML shows that the lack of run-time efficiencies and scalability is an insignificant source of concern at research levels.

Figure 3.3: Example of a Streamlit Application

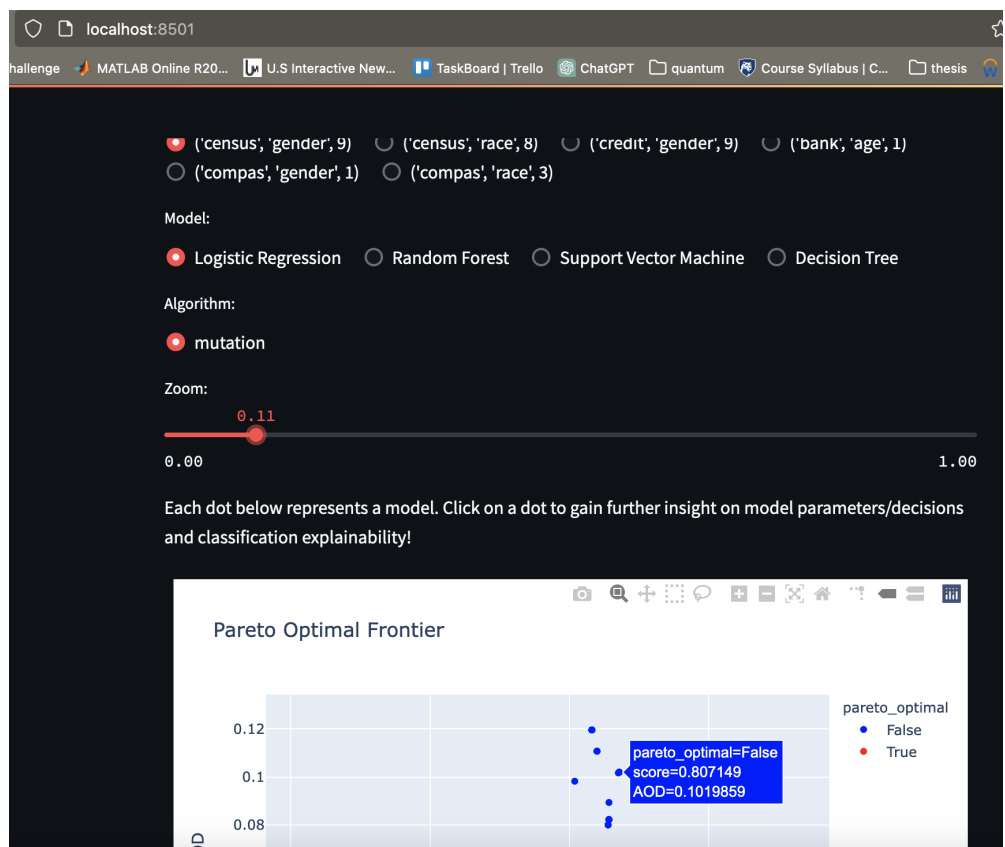

Streamlit apps create a web server on your local machine, leveraging your browser to act as a GUI to communicate with your simple Python scripts.

### 3.3.2 Providing Human-Centric Design with Plotly

Plotly integrates seamlessly with Streamlit to display basic statistical results such as bar graphs, scatterplots, pie charts, and more [17]. Furthermore, Plotly provides a seamless GUI for each graph to enable a more human-centric infrastructure. In many cases, displaying all of the data or analysis to the user is both distracting and computationally infeasible in real time. Therefore, we generate an ambiguous summary for each option using a few metrics, plot them using Plotly, and allow users to click directly on the graph to see more details about each option. Figure 3.4 provides some examples of this philosophy. In the case of FairLay-ML's data visualization tool, a summary of correlations between each feature in the dataset to the sensitive feature is first provided; when a user clicks on a specific feature, additional details in the form of a histogram provide distribution discrepancies for each feature conditioned on different sensitive categories, as seen in Figure 3.4a (see Section 3.4 for a specific example). Since Parfait-ML generates a large number of models with varying fairness and accuracies, we summarize each model in a scatterplot by plotting each model's accuracy score and AOD score; as shown in Figure 3.4b, users can then click on each point to see the logic[1] behind each model. In specific, Figure 3.4b shows the logistic regression models Parfait-ML trained for race fairness using the Census data set by varying hyperparameters. The scatterplot plots each model according to its AOD score and accuracy. By clicking on a point, the specific model's logic (as well as other information about the model) is displayed. For this specific case, the logic is represented by the weights used for the weighted sum in logistic regression models. Finally, Figure 3.4c demonstrates the scatterplot of each training data point's predicted probability calculated by a chosen model, and LIME's explanation for the prediction probability is shown to users when they click on a training data point. Figure 3.4c specifically shows a logistic regression model trained for fairness against race using the Census dataset. Each point represents a data point within the training dataset, with the red points indicating counterfactuals. As hypothesized in Section 2.3, most counterfactuals lie near the predicted probability of 0.5. When we click on a specific data point, a LIME plot shows how the feature values of the data point impact the data point's prediction probability value.

---

[1]FairLay-ML's display of model logic is summarized in Figure 1.1 under Section 1.2

Figure 3.4: Simplified Plotly Graphs Interactive Display More Details On-Demand

(a) Correlation Simplification for Sensitivity Histograms Study

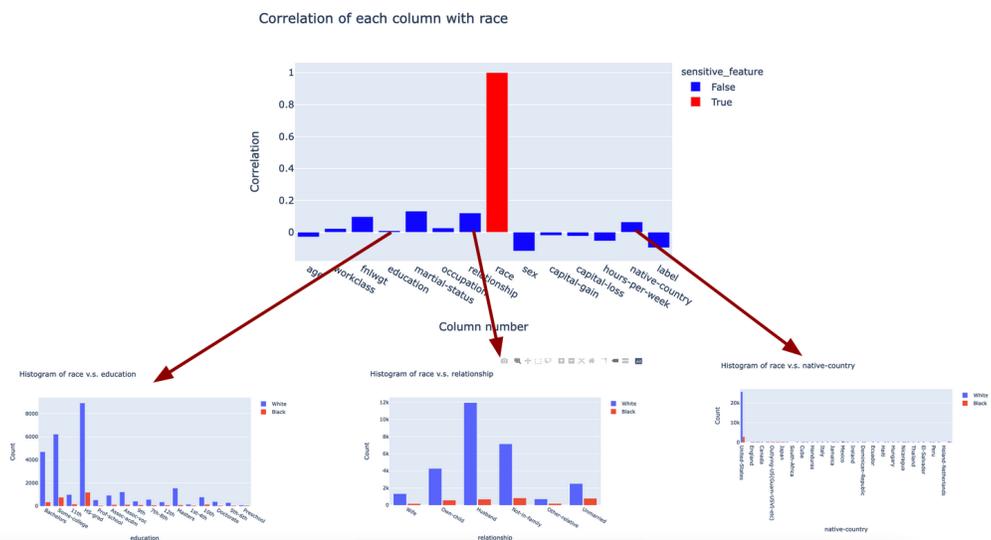

(b) Accuracy and Fairness Scores Simplification for Model Logic

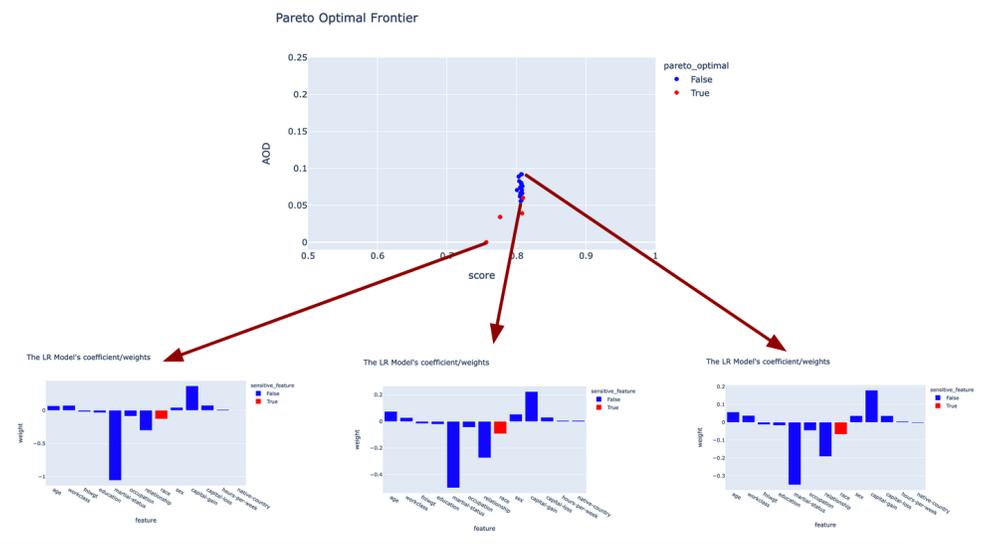

(c) Predicted Probability Simplification for LIME Explanations

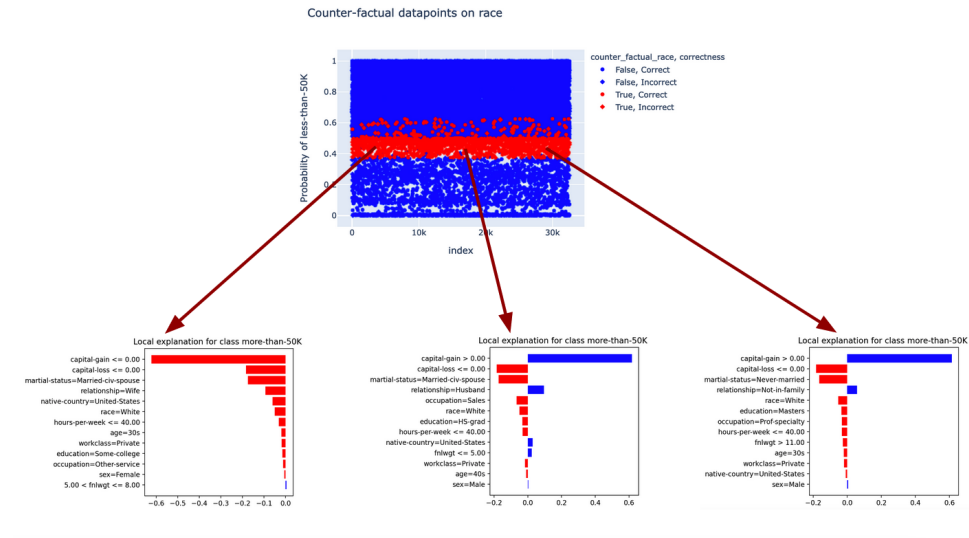

Plotly graph is utilized to summarize the details of data points, models, or features. Users can click on the graphs directly to request additional details on clicked points. This allows users to access all available information without overwhelming the user.

## 3.4 Feature Visualization

Leveraging Parfait-ML's labeled datasets as discussed in Section 3.2 and tools discussed in Section 3.3 to ingest the data files, we design a data visualization tool that can provide insights regarding how a sensitive attribute is tied to other features. Correlation for each feature against the sensitive attribute is first plotted. For each case-study dataset, a Plotly bar graph is generated in real time calculating the correlation between each feature with the sensitive attribute using Pandas. The Plotly graph doubles as an intuitive GUI to allow users to quickly choose a specific feature column of interest and compare it against the sensitive feature column. As shown in Figure 3.4a, additional details show up when you click on the correlation bar graph for the specific feature chosen. By comparing just one feature against the sensitive feature, we can provide significantly more in-depth data without overwhelming the user. For categorical features, the occurrence of each user in each category is split into two bars – one for users in the category which is also in the first sensitive category (e.g. male for gender), and the other for users in the category which is also in the second sensitive category (e.g. female for gender)[2]. We bin numerical features and perform a similar operation.

Figure 3.4a specifically demonstrates a correlation between features in the Census dataset against race. The red bar indicates the sensitive attribute itself. This "dummy check" shows that the correlation of the sensitive attribute with itself is 1. We can see that native country does not have a specifically high correlation metric in the summary. As explained above, this is because the numerical encoding of the countries is arbitrary. The histogram showing the number of data points in

---

[2]For simplicity in our study, sensitive attributes only have two categories.

each category that falls into sensitive attribute subcategories is shown on the right graph. However, in this case, we have what is called a "long-tailed" distribution, with the most data points coalescing in the "United States" category. This makes the analysis of other countries more difficult to analyze.

Hence, we also provide an additional graph, which specifies the percentage of data points in each category in a feature that is in sensitive category 1. While this provides the same information, this graph better highlights the categories within a feature (i.e. feature with percentages very far away from the average percentage) that may contribute to a model's unfairness. The equivalent graph of the native-country histogram explained above is shown later in Figure 4.4.

## 3.5  Visualizing Models

In Section 3.2.1, we provide the updates needed to save the model logic of each trained machine learning model into a pickle file. This file format is meant to be re-ingested by the same Python pickle library to recreate the original Python scikit-learn objects. For Logistic Regression and Support Vector Machine models, we again utilize Plotly to plot the weights that the models attribute to. We understand that these weights may be misleading, as a small weight on a feature column can have drastic impacts on the final result if the feature column has a high variance. Therefore, we generate two graphs: one with raw weights and one with weights adjusted for the feature column's standard deviation. For Decision Trees and Random Forests, a scikit-learn library is available to generate a visualization of the decision tree logic. Some Random Forest models come with more than 50 to 80 trees; we allow users to choose which models they want to visualize. Figure 1.1 in the Background section provides in-depth explanations for how each model is represented.

## 3.6  Explaining Models with LIME

We pick a blackbox explainability package called Local Interpretable Model-Agnostic Explanation, or LIME, to explain why a model predicts a data point [18]. Given a data point, this package provides an intuitive visualization of the importance of each feature to the final classification. To do this, LIME takes a machine learning model as a blackbox function with inputs equivalent to the feature space and output, along with the model's confidence in the possible outputs. LIME then perturbs different features from the original data point and observes for changes in the model's confidence in the possible outputs to determine how much each feature impacts the decision.

As the name suggests, LIME is model-agnostic, meaning that it is built to explain any classification black-box algorithm. It is compatible with more modern classification tools such as deep neural networks, which are fundamentally more difficult to visualize any other way. Furthermore, it is equipped to explain various data sources, including high dimensional datasets such as imagery datasets and natural language datasets, and provide intuitive graphics. These extensions can be used to explain unfairness in computer vision and natural language processing tools.

In Section 3.2.4, we discuss custom slight changes to Parfait-ML that enforced scikit-learn models to provide a probability or confidence score when sampled. This laid the groundwork for us to create a custom Python Lambda function base on the scikit-learn models to feed to LIME, similar to the lambda function we fed to Themis.

### 3.6.1  Labeling Explainability

In Section 2.2, we laid the groundwork for picking a model and an individual point for an unfairness explanation. Specifically, users can pick any point from the data set to be shown explanations for how important the model interprets each feature at its given position. In Section 3.2.2, we note that we encode the meaning of each feature directly on the dataset. This encoding automatically gets pushed through Pandas when we pull the dataset and into LIME. However, for categorial features, we create a custom API to extract the meaning of the numerical encoding for the meaning of each point's position in the feature space. This allows LIME to provide meaning for the local environment in which the data point is sampled when testing and visualizing explainability.

### 3.6.2  Allowing User to Explain Any Point in Feature Space

We later added functionality to allow users to create any point in the feature space for measurements. Using standard Streamlit input interfaces, users pick from a list of possible values for categorical features and type in numerical values for numerical features. While less true for more classical models we are analyzing as compared to deep models (an example being convolution), certain combinations could carry heavy weights collectively but at other positions, the feature may not hold any weight at all.

### 3.6.3  Creating a Data-Invariant View of Model Explainability Through LIME

We primarily study counterfactuals during our explainability study as they are most impacted by their sensitive attribute category. If no specific data point is picked, we create a random sample from counterfactuals and also a sample of random actual data points. Using weights provided by LIME, we then calculate the average of the absolute importance weights for each feature to generate a more holistic understanding of which features are overall most important. This method has the disadvantage that the probability impact of each feature is absolute-averaged among the categories in the feature. When a data point is picked, additional details provide the exact category that is impacting prediction probability.

# Chapter 4

# Results

# 4.1   Models Explained

As a proof-of-concept, we take the least fair (i.e. lowest AOD score) model generated by Parfait-ML. These models provide examples of remedies that are statistically significant and clear. In practice, users are more likely to use FairLay-ML on their most accurate models, which may have various fairness metrics depending on the situation. In such cases, FairLay-ML may provide intuitions with less confidence.

## 4.1.1   Some Specific Examples

The simplest example is when unfairness stems from the sensitive attribute itself. Figure 4.1 provides one such scenario. In this case, age itself is heavily weighted in the logistic regression model, and the use of LIME reinforces that model decision is indeed impacted by the age feature.

The integration of LIME demonstrates its value as the logic becomes more complex. Figure 4.2 and Figure 4.3 provide intuitive examples where LIME demonstrates its value. By testing the model on specific data points, LIME determines how specific categories within a feature change prediction probabilities. In this case, the user can reach the conclusion easily: the relationship feature, which this model values in its decision-making process, accidentally encoded gender information. However, the relationship feature has other categories which do not heavily contribute to unfairness, such as "own children", "unmarried", and "other relatives". Without LIME, these categories which do not impact fairness may obfuscate a user's ability to see that the relationship feature is impacting fairness. Furthermore for the case of decision trees in Figure 4.3, ranking which feature is making an impact on the final classification is difficult to measure without LIME from the start. In this case, instead of categorizing someone as "husband" and "wife", it may be wise for the user to combine the two categories to just "married". We use this intuition in Section 4.1.2.

Some examples are even less obvious. In some cases, the sensitive feature may be correlated with another feature in a way that may only be obvious to domain experts. In Figure 4.4, we provide one such less intuitive example, where the native country of a user provides a strong indication of their race. Our data visualization tool comes in handy in these scenarios. Our histogram on the dataset is able to quickly inform us that the native country, which our model discriminates against, can provide a strong indication of someone's race. In this way, even data scientists without domain expertise (for this specific case, in each country's demographics) can provide logical explanations for why a model is unfair.

Figure 4.1: Simplest Example of LIME Explanation

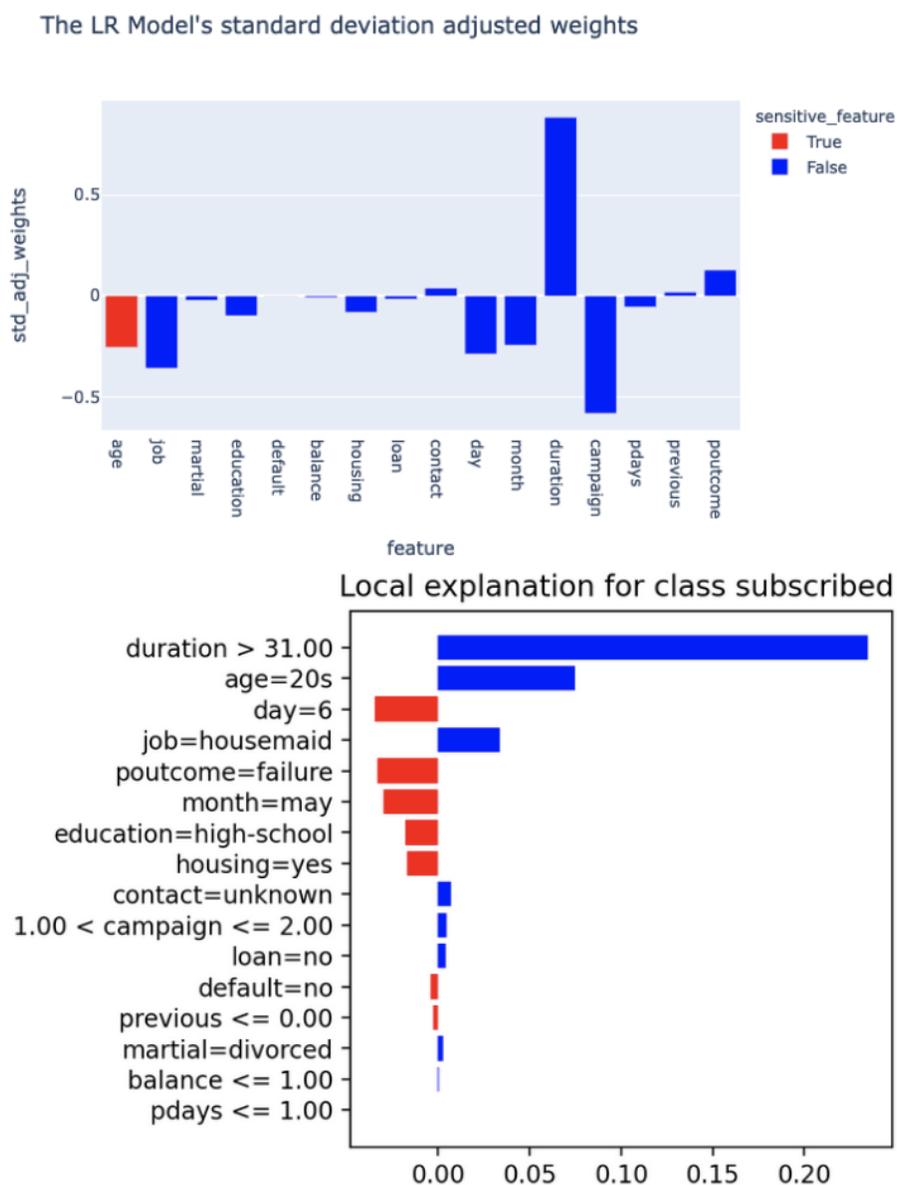

The top image shows the weights of each feature divided by the sample deviation of each feature to show the relative importance of the feature to the model on the most unfair Logistic Regression model with respect to age trained on the Bank dataset. The bottom image demonstrates LIME's black box explanation for a counterfactual on the model. In this case, the sensitive attribute itself (age) is the second most impactful to decision probability.

Figure 4.2: Intuitive Example of LIME Providing Valuable Insight

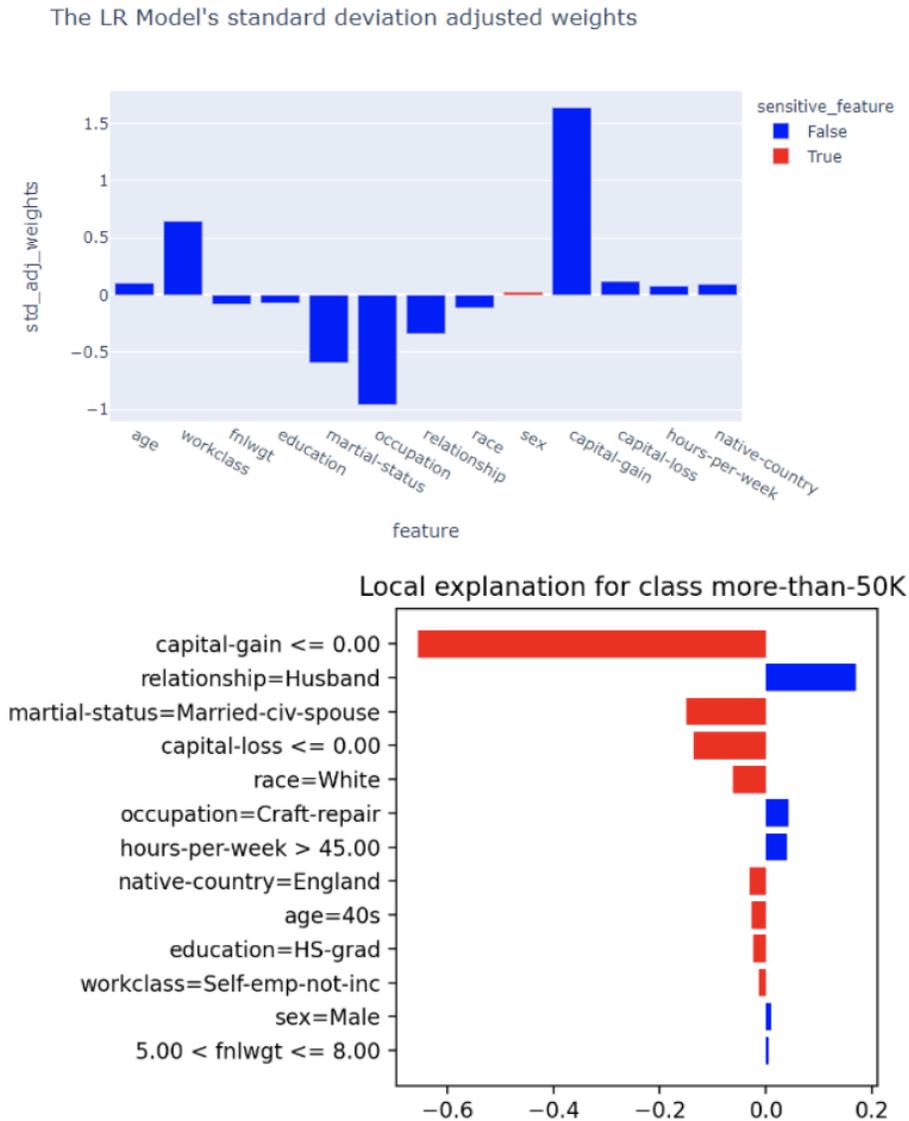

In this case, our model logic graph on the most unfair Logistic Regression model with respect to gender trained on the Census dataset (top) does not immediately provide insights regarding why the model is unfair. However, the LIME explanation (bottom) for the model provides intuition: while gender itself does not strongly influence decision probability, we see that the relationship status of "husband" – which encodes the gender information – is the second most influential to model decision probabilities.

Figure 4.3: Intuitive Example of LIME Providing Valuable Insight 2

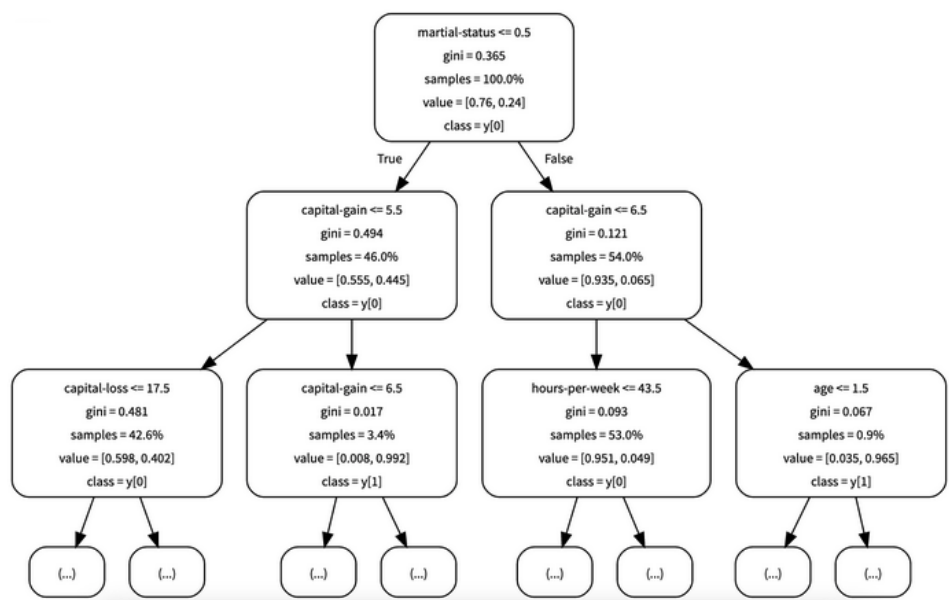

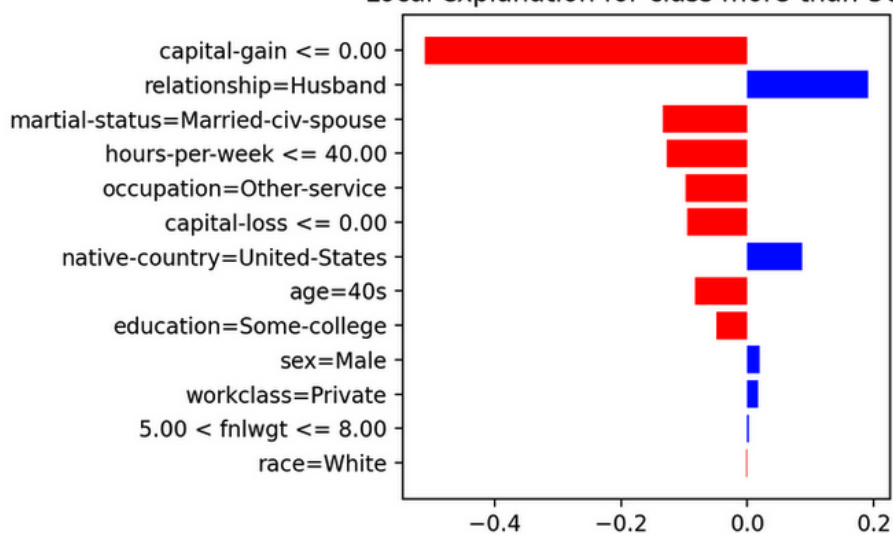

Similar to our first intuitive example, our model logic graph on the most unfair Decision Tree model with respect to gender trained on the Census dataset (top) does not immediately provide insights regarding why the model is unfair. However, the LIME explanation (bottom) for the model provides intuition: while gender itself does not strongly influence decision probability, we see that the relationship status of "husband" – which encodes the gender information – is again the second most influential to model decision probabilities.

Figure 4.4: Less Intuitive Example of LIME Providing Valuable Insight

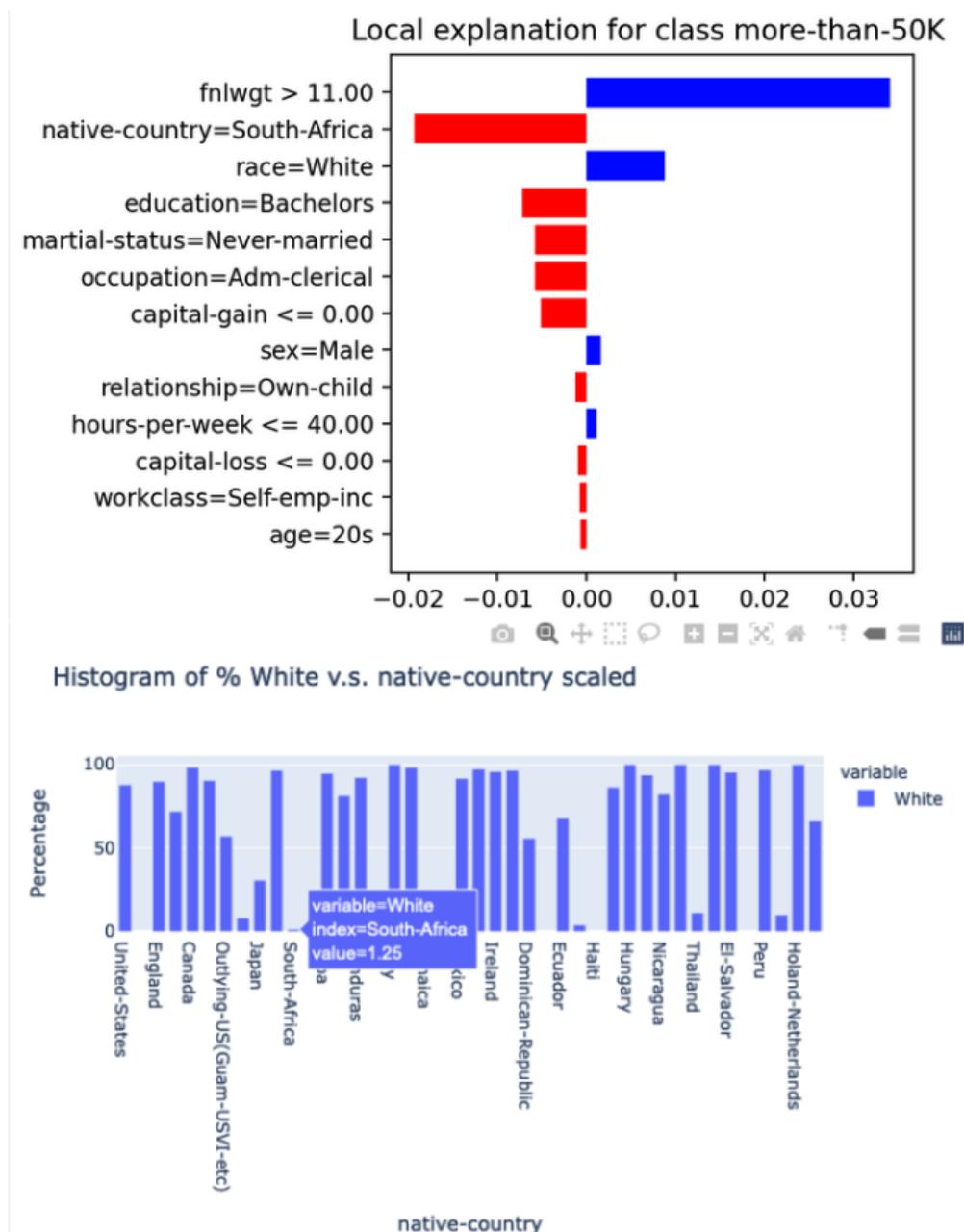

This LIME explanation (top) is provided for a counterfactual on the most un-fair Support Vector Machine model with respect to race trained on the Census dataset. We see that in this case, someone's native country is the second most impactful to model decisions. The histogram (bottom) demonstrates that a person from South Africa has a very low probability of being White, indicating that this category (South Africa) can be used by the model to determine someone's race and cause bias.

### 4.1.2 Remedies to Unfairness

These explanations for how bias and correlation in the data affect bias in the logic of the models help provide a remedy to the models. One intuitive remedy is to mask the features that the model uses for discrimination against a sensitive group. This is a more targeted version of "fairness through unawareness", which has been demonstrated to be ineffective precisely because other features can be correlated to the sensitive attribute [14]. We confirm this result using our work in Section 3.2.3. Indeed, in 3 of our 4 examples, our remedy does not involve masking the sensitive attribute at all but rather an attribute correlated to it:

- On the least fair logistic regression (LR) model on age for the Bank dataset, we mask the age feature, as it is strongly impacting model decisions, causing unfairness. In order to maintain the architecture of the model (as changing the number of inputs could change the meaning of the hyperparameters, which we keep constant), we simply mask the age column to the same value across all data points.

- On both the least fair logistic regression (LR) and least fair decision tree (DT) model on gender for the Census dataset, we mask the relationship of wife and husband, as the relationship is strongly impacting model decisions. Obviously, wife encodes female gender and husband encodes male gender [1]. The other categories, including "own children", "unmarried", and "other relatives", are untouched as this information does not correlate to gender as strongly and may be helpful to model accuracy.

- On the least fair support vector machine (SVM) model on race for Census, we mask the native country, which is again strongly impacting model decisions. Since the native country is a categorical feature, the correlation metric is artificially low because countries with low white populations are arbitrarily mixed in with countries with high white populations during numerical encoding. However, a histogram of the white population (Figure 4.4) in various native countries provide a very clear picture: some native countries have significantly more white population than others.

Table 4.1 provides the results comparing the originally trained models against the models trained under the masked dataset. We clearly see that the accuracy (score) is very similar in all three cases, while the AOD score decreases significantly for three of the models. For the least fair decision tree model on gender for Census, there is only a slight decrease in AOD. This demonstrates that in most of our cases, the intuitive remedies derived from the data successfully hide the bias in the data leveraged by unfair models without hiding crucial information necessary to make an accurate prediction.

The unremedied and remedied models are also tested with Themis, providing results seen in Table 4.2. The results, based on a newly generated set of feature data, concur with conclusions drawn from Parfait-ML's AOD scores.

The difference between the unremedied and remedied models' Themis causal fairness metrics appears to contradict this conclusion. We highlight some potential reasons for this in Section 4.3.1.

Overall, we provide intuitive explanations and accompanying remedies for 4 of the least fair models out of the 24 case studies. As shown in Table 4.1 and Table 4.2, three of the four mod-

---

[1]A dummy check can be performed on the data visualization tool to verify.

Table 4.1: Comparing Accuracy and AOD for Unremedied and Remedied Models

| Model | Unmasked Score | Masked Score | Unmasked AOD | Masked AOD |
|---|---|---|---|---|
| Least Fair Age LR for Bank | .8817 | .8886 | 0.07719 | 0.02968 |
| Least Fair Gender LR for Census | .8032 | .8037 | .1195 | .05730 |
| Least Fair Gender DT for Census | .8020 | .8052 | .06953 | .06733 |
| Least Fair Race SVM for Census | .6988 | .6441 | .1689 | .02618 |

The exact strategy for masking is detailed in Section 4.1.2. This table compares the original (unremedied/unmasked) models compared to the remedied models trained on based data.

Table 4.2: Comparing Themis Score For Rememedied and Unremedied Models

| Model | Unmasked group | Masked group | Unmasked causal | Masked causal |
|---|---|---|---|---|
| Least Fair Age LR for Bank | .08643 | .01961 | .07071 | 0 |
| Least Fair Gender LR for Census | .006371 | .001217 | 0 | 0.01379 |
| Least Fair Gender DT for Census | .01765 | .01608 | 0 | .009709 |
| Least Fair Race SVM for Census | .05088 | .02038 | .01695 | .02602 |

The exact strategy for masking is detailed in Section 4.1.2. This table compares the original (unremedied/unmasked) models compared to the remedied models trained on mased data.

els perform significantly more fairly in terms of group discrimination after remedy. The fourth performs negligibly more fairly.

## 4.2 Infrastructure Summary

Our deployment of FairLay-ML's software stack on `https://fairlay-ml.streamlit.app/` uses only free computational resources, demonstrating that our explainability pipeline works in near real-time scale with reasonable computational power. Our GitHub repo (`https://github.com/Pennswood/FairLay-ML/tree/live-streamlit-app`), where the code for FairLay-ML's software stack is hosted, provides the necessary packages to install: Python, pip, and eight lightweight packages through pip specified in "requirements.txt".

This demonstrates the successes of our core philosophies and components:

1. Summarizing details onto a Plotly graph and computing additional details as users request them interactively by clicking on the graphs provides two significant benefits: users are less overwhelmed and gain more clarity while allowing a significant decrease in necessary computation.

2. Many of our components, including LIME and generating a visualization of model logic, can be computed in real-time for the user, decreasing the needed memory to store them in our servers.

3. Overall, free and low-cost infrastructures exist (Streamlit.io, Heroku, etc) to rapidly deploy and share data visualization infrastructure from GitHub.

Ultimately, our infrastructure achieves our core goal to help humans understand the unfairness in machine learning models and datasets on an intuitive level. Even more so, we demonstrate that this understanding can translate to smarter, more informed decisions on how to best mitigate unfairness.

## 4.3 Possible Concerns

### 4.3.1 Addressing Themis' Shortcomings

The primary benefit of Themis lies in its ability to test models for fairness using its own randomly generated data of the feature space, without involvement from our limited training dataset. However, from the outset, there are two clear concerns. First, Themis generates data by uniform sampling from the feature space, a distribution that is unlikely to be truly representative of the actual feature distribution. Second, Themis must use an alternative definition of unfairness because its own generated data is unlabeled. We attempted to alleviate this concern by running Parfait-ML generated models against the Themis test. We clearly see a positive correlation between both of the tested fairness scores of Themis compared to AOD scores provided by Parfait-ML (see Table 4.3).

**Themis' Group Discrimination Score**

In 20 of the 24 studies, the Themis group discrimination score is higher for most unfair compared to most fair (see Table A.1. The cause for the 4 studies where the most unfair model outperforms the most fair model is different for each scenario. Randomness from the Themis' random sampling contributes to the group Themis scores. Furthermore, Themis samples randomly from the whole feature space, a method that is not indicative of the true population distribution.

This result demonstrates that Themis' group discrimination score overall concurs with Parfait-ML's AOD score.

**Themis' Causal Discrimination Score**

The case for Themis' causal discrimination score is less strong. As shown in Table 4.4, Themis is more likely to find no counterfactuals at all, providing a discrimination score of 0, even when the training data clearly demonstrate that certain locations of the feature space contain counterfactuals. This indicates that a uniform random distribution is not an accurate model for counterfactuals fairness analysis. Still, the overall trends agree. As seen in Table 4.4, both Parfait-ML and Themis are most likely to find no counterfactuals at all for the "most fair" models in the dataset and through random sampling respectively, with "most accurate" coming second and "most unfair" coming last.

Table 4.3: Themis Discrimination Score Summary

| Metric | Most Unfair | Most Accurate | Most Fair |
|---|---|---|---|
| Avg. Odd Diff. | .09468 | .031318 | .012054 |
| Themis group score | .055052 | 0.025369 | 0.010326 |
| Themis causal score | 0.09468 | 0.03132 | 0.01205 |

Across our 6 scenarios times 4 model-type each for a total of 24 studies, the Themis and Parfait-ML fairness metrics are averaged for the most fair, most accurate, and most unfair models. For each group metric test, we sample 50,000 random points from each sensitive group for the group metric. For each causal metric test, a total of 50,000 random samples from the feature space are taken. Each model's scores can be found in Table A.1.

Table 4.4: Comparing Counterfactuals (cf) Found in Themis and Parfait-ML

| Num. studies with no cf | Most Unfair | Most Accurate | Most Fair | Total |
|---|---|---|---|---|
| Parfait-ML | 7 | 11 | 17 | 35 |
| Themis | 12 | 16 | 22 | 50 |

Parfait-ML tests its model for counterfactuals based on a model with the training datasets ranging in size from 1000 to 45,211. Themis tests Parfait-ML's models using 50,000 uniform, randomly generated points from the feature space. Counterfactuals are computed with respect to the sensitive feature of the case-study. This table summarizes the number of models in which Parfait-ML and Themis found no counterfactuals at all, out of a total of 72 models from 24 case studies in each category of "most fair", "most accurate", and "most unfair".

## 4.3.2 Cherry-Picked Models

Through our investigation of the models, we find four models in which we provide possible explanations and remedies out of the twenty-four test cases[2]. Of the four models, three provide significantly lower group fairness scores after remedy.

The reason for the low number of models varies. For the two Compas dataset studies on race and gender, we are unable to identify what each feature column represents. Therefore, intuitive examples are difficult for the 8 test cases involving the Compas dataset. For the cases of the least fair decision trees w.r.t. race for Census and gender for Credit, the model consisted of one node. In other words, the decision tree model makes the same prediction for all data points. Therefore, the discussion of fairness in these cases is meaningless. The probability predictions of Support Vector Machines and Random Forests are also overall skewed. Since LIME utilizes the probability of prediction to provide an explanation, this likely obfuscated some of LIME's explainability power for those models.

---

[2]Each test case comes with 3 models. However, in the interest of demonstration, we pick the least-fair model for analysis

These issues compounded, leading to only four models in which LIME and FarLay-ML's data visualization tool provided a clear, intuitive explanation for their unfairness.

## 4.4   Future Study

One clear extension to the current architecture is the completion of the MLOps cycle. Currently, FairLay-ML provides the necessary components for explaining unfair models, as demonstrated by the orange boxes in Figure 2.1. In the future, we envision FairLay-ML's infrastructure to encompass updating the dataset and retraining models directly through the GUI. In particular, users will be able to mask different feature columns (or just categories within the feature), retrain the model, and immediately view the updated model through our GUI for accuracy and fairness metric changes.

As our contradicting results between Themis' group score and Themis' causal score suggests, a more in-depth analysis of the connection between the group score and the causal score is needed. This analysis may also provide further directions regarding how group fairness and counterfactual fairness may be related, and under what circumstances they may diverge.

We would also like to incorporate more in-depth studies on how the hyperparameter directly impacts the logic of the model, which in turn impacts the fairness of the model. Using LIME and the data visualization tool, an analysis can be performed on models with different hyperparameters regarding how each model perpetrates unfairness. This may shine a light on how the hyperparameter impacting the model logic further causes unfairness downstream.

As real-world use cases for machine learning models become more complex, it will also become very important to analyze how our infrastructure scales with significant increases in data dimensions. LIME has integration capabilities for data types such as images and sentences. However, in-depth investigations will need to be made regarding data visualizations for images and audio.

# Appendix A

# Models Accuracy and Fairness

Table A.1: Original Models Accuracy and Fairness Scores

| score | AOD | group_score[1] | causal_score[2] | dataset | model | optimal[3] |
|---|---|---|---|---|---|---|
| 0.8032 | 0.1195 | 0.0064 | 0 | census, gender | LR | Worst |
| 0.8085 | 0.0391 | 0.0004 | 0.0097 | census, gender | LR | Score |
| 0.8071 | 0.0169 | 0.0052 | 0 | census, gender | LR | Fairness |
| 0.835 | 0.1861 | 0.207 | 0.2032 | census, gender | RF | Worst |
| 0.8543 | 0.0949 | 0.0033 | 0 | census, gender | RF | Score |
| 0.7654 | 0.0347 | 0 | 0 | census, gender | RF | Fairness |
| 0.7097 | 0.1249 | 0.0017 | 0.0169 | census, gender | SV | Worst |
| 0.7192 | 0.003 | 0.0018 | 0 | census, gender | SV | Score |
| 0.7192 | 0.003 | 0.0018 | 0 | census, gender | SV | Fairness |
| 0.802 | 0.0695 | 0.0177 | 0 | census, gender | DT | Worst |
| 0.802 | 0.0695 | 0.0177 | 0 | census, gender | DT | Score |
| 0.7565 | 0 | 0 | 0 | census, gender | DT | Fairness |
| 0.8071 | 0.092 | 0.0518 | 0.0457 | census, race | LR | Worst |
| 0.8094 | 0.06 | 0.0245 | 0 | census, race | LR | Score |
| 0.7565 | 0 | 0 | 0 | census, race | LR | Fairness |
| 0.803 | 0.0975 | 0.026 | 0.0097 | census, race | RF | Worst |
| 0.8552 | 0.072 | 0.0184 | 0 | census, race | RF | Score |
| 0.8509 | 0.0489 | 0.0715 | 0 | census, race | RF | Fairness |
| 0.6988 | 0.1689 | 0.0509 | 0.0169 | census, race | SV | Worst |
| 0.7215 | 0.0216 | 0.0464 | 0.0445 | census, race | SV | Score |
| 0.2436 | 0.0008 | 0 | 0 | census, race | SV | Fairness |
| 0.8243 | 0.0899 | 0.0204 | 0.0097 | census, race | DT | Worst |
| 0.8522 | 0.0451 | 0.0234 | 0 | census, race | DT | Score |
| 0.7565 | 0 | 0 | 0 | census, race | DT | Fairness |
| 0.748 | 0.1102 | 0.0821 | 0.075 | credit, gender | LR | Worst |
| 0.772 | 0.0336 | 0.0723 | 0.0652 | credit, gender | LR | Score |
| 0.704 | 0 | 0 | 0 | credit, gender | LR | Fairness |
| 0.748 | 0.1113 | 0.0086 | 0 | credit, gender | RF | Worst |
| 0.772 | 0.017 | 0.0132 | 0.041 | credit, gender | RF | Score |
| 0.704 | 0 | 0.0041 | 0 | credit, gender | RF | Fairness |
| 0.672 | 0.0852 | 0.0249 | 0 | credit, gender | SV | Worst |
| 0.672 | 0.0852 | 0.0249 | 0 | credit, gender | SV | Score |
| 0.328 | 0.0718 | 0.0278 | 0 | credit, gender | SV | Fairness |
| 0.7 | 0.1797 | 0.0036 | 0 | credit, gender | DT | Worst |
| 0.74 | 0.0374 | 0.0739 | 0.0799 | credit, gender | DT | Score |
| 0.704 | 0 | 0 | 0 | credit, gender | DT | Fairness |
| 0.8817 | 0.0772 | 0.0864 | 0.0707 | bank, age | LR | Worst |

[1]Themis

[2]Themis

[3]Worst optimal means least fair; score means most accurate; fairness means most fair.

| 0.8955 | 0.0063 | 0.0323 | 0.0313 | bank, age | LR | Score |
| 0.8887 | 0.002 | 0.0169 | 0.0097 | bank, age | LR | Fairness |
| 0.8817 | 0.0428 | 0.0229 | 0 | bank, age | RF | Worst |
| 0.9014 | 0.0109 | 0.1443 | 0.2717 | bank, age | RF | Score |
| 0.8828 | 0 | 0 | 0 | bank, age | RF | Fairness |
| 0.4453 | 0.1393 | 0.3217 | 0.3295 | bank, age | SV | Worst |
| 0.8753 | 0.0306 | 0.0483 | 0.0422 | bank, age | SV | Score |
| 0.6369 | 0.0174 | 0.0726 | 0.0901 | bank, age | SV | Fairness |
| 0.8713 | 0.0289 | 0.0559 | 0.2432 | bank, age | DT | Worst |
| 0.8828 | 0 | 0 | 0 | bank, age | DT | Score |
| 0.8828 | 0 | 0 | 0 | bank, age | DT | Fairness |
| 0.9152 | 0.1374 | 0.1244 | 0.1517 | compas, gender | LR | Worst |
| 0.9706 | 0.0163 | 0.0135 | 0 | compas, gender | LR | Score |
| 0.9706 | 0.0163 | 0.0135 | 0 | compas, gender | LR | Fairness |
| 0.964 | 0.0284 | 0.0139 | 0 | compas, gender | RF | Worst |
| 0.9706 | 0.0163 | 0.0135 | 0 | compas, gender | RF | Score |
| 0.9706 | 0.0163 | 0.0135 | 0 | compas, gender | RF | Fairness |
| 0.9706 | 0.0163 | 0.0135 | 0 | compas, gender | SV | Worst |
| 0.9706 | 0.0163 | 0.0135 | 0 | compas, gender | SV | Score |
| 0.9706 | 0.0163 | 0.0135 | 0 | compas, gender | SV | Fairness |
| 0.6319 | 0.1146 | 0.0022 | 0 | compas, gender | DT | Worst |
| 0.9706 | 0.0163 | 0.0135 | 0 | compas, gender | DT | Score |
| 0.5471 | 0 | 0 | 0 | compas, gender | DT | Fairness |
| 0.9152 | 0.1687 | 0.1554 | 0.1456 | compas, race | LR | Worst |
| 0.9706 | 0.015 | 0.0025 | 0 | compas, race | LR | Score |
| 0.9706 | 0.015 | 0.0025 | 0 | compas, race | LR | Fairness |
| 0.964 | 0.0216 | 0.0011 | 0 | compas, race | RF | Worst |
| 0.9706 | 0.015 | 0.0025 | 0 | compas, race | RF | Score |
| 0.9706 | 0.015 | 0.0025 | 0 | compas, race | RF | Fairness |
| 0.9706 | 0.015 | 0.0025 | 0 | compas, race | SV | Worst |
| 0.9706 | 0.015 | 0.0025 | 0 | compas, race | SV | Score |
| 0.9706 | 0.015 | 0.0025 | 0 | compas, race | SV | Fairness |
| 0.847 | 0.0475 | 0.0204 | 0 | compas, race | DT | Worst |
| 0.9706 | 0.015 | 0.0025 | 0 | compas, race | DT | Score |
| 0.5471 | 0 | 0 | 0 | compas, race | DT | Fairness |

Table A.2: Remedied Models Accuracy and Fairness Scores

| score | AOD | group_score[4] | causal_score[5] | dataset | model | optimal[6] |
|---|---|---|---|---|---|---|

---

[4]themis

[5]themis

[6]Worst optimal means least fair; score means most accurate; fairness means most fair.

| | | | | | | |
|---|---|---|---|---|---|---|
| 0.8037 | 0.0573 | 0.0012 | 0.0138 | census, gender | LR | Worst |
| 0.8067 | 0.0534 | 0.0155 | 0 | census, gender | LR | Score |
| 0.8004 | 0.0343 | 0.0062 | 0 | census, gender | LR | Fairness |
| 0.8374 | 0.1625 | 0.1268 | 0.1054 | census, gender | RF | Worst |
| 0.8569 | 0.0979 | 0.0114 | 0 | census, gender | RF | Score |
| 0.7666 | 0.0473 | 0 | 0 | census, gender | RF | Fairness |
| 0.7132 | 0.0356 | 0.0074 | 0.0097 | census, gender | SV | Worst |
| 0.7223 | 0.0063 | 0.0034 | 0 | census, gender | SV | Score |
| 0.7223 | 0.0063 | 0.0034 | 0 | census, gender | SV | Fairness |
| 0.8052 | 0.0673 | 0.0161 | 0.0097 | census, gender | DT | Worst |
| 0.8052 | 0.0673 | 0.0161 | 0.0097 | census, gender | DT | Score |
| 0.7565 | 0 | 0 | 0 | census, gender | DT | Fairness |
| 0.8079 | 0.091 | 0.0534 | 0.0422 | census, race | LR | Worst |
| 0.809 | 0.0632 | 0.035 | 0 | census, race | LR | Score |
| 0.7565 | 0 | 0 | 0 | census, race | LR | Fairness |
| 0.8015 | 0.1034 | 0.0138 | 0 | census, race | RF | Worst |
| 0.8544 | 0.0733 | 0.0171 | 0 | census, race | RF | Score |
| 0.8514 | 0.0377 | 0.0702 | 0 | census, race | RF | Fairness |
| 0.6441 | 0.0262 | 0.0204 | 0.026 | census, race | SV | Worst |
| 0.7129 | 0.01 | 0.0138 | 0 | census, race | SV | Score |
| 0.7129 | 0.01 | 0.0138 | 0 | census, race | SV | Fairness |
| 0.8399 | 0.0459 | 0.018 | 0 | census, race | DT | Worst |
| 0.8526 | 0.0405 | 0.0235 | 0.0097 | census, race | DT | Score |
| 0.7565 | 0 | 0 | 0 | census, race | DT | Fairness |
| 0.8886 | 0.0297 | 0.0196 | 0 | bank, age | LR | Worst |
| 0.8955 | 0.0123 | 0.0282 | 0 | bank, age | LR | Score |
| 0.8955 | 0.0123 | 0.0282 | 0 | bank, age | LR | Fairness |
| 0.8817 | 0.0428 | 0.0229 | 0 | bank, age | RF | Worst |
| 0.9001 | 0.0101 | 0.032 | 0 | bank, age | RF | Score |
| 0.8828 | 0 | 0 | 0 | bank, age | RF | Fairness |
| 0.7909 | 0.0536 | 0.0368 | 0 | bank, age | SV | Worst |
| 0.8396 | 0.0327 | 0.0214 | 0 | bank, age | SV | Score |
| 0.8396 | 0.0327 | 0.0214 | 0 | bank, age | SV | Fairness |
| 0.8703 | 0.0077 | 0.0271 | 0 | bank, age | DT | Worst |
| 0.8828 | 0 | 0 | 0 | bank, age | DT | Score |
| 0.8828 | 0 | 0 | 0 | bank, age | DT | Fairness |

We only study the least fair models on some select model type for each dataset with fairness w.r.t to certain sensitive attributes. However, our algorithm generates remedied models for all of the models. The raw results for all of the models are detailed in this table.

# Appendix B

# Hyperparameter Used for Each Model

Table B.1: Decision Tree Model Hyperparameters

| dataset | optimal | criterion | splitter | max.depth | min.samples.split | min.samples.leaf | min.weight.fraction.leaf | max.features | random.state | max.leaf.nodes | min.impurity.decrease | class.weight | ccp.alpha |
|---|---|---|---|---|---|---|---|---|---|---|---|---|---|
| census, gender | Fairness | entropy | random | None | 3 | 5 | 0.329706872 | None | 2019 | None | 0 | None | 0.0289456813 |
| census, race | Fairness | gini | best | 18 | 3 | 2 | 0.1371628281 | log2 | 2019 | None | 0 | None | 0.1569021191 |
| credit, gender | Fairness | entropy | random | 5 | 5 | 3 | 0.2384499133 | sqrt | 2019 | None | 0 | None | 0.9071772715 |
| bank, age | Fairness | gini | best | None | 2 | 1 | 0.446090821 | None | 2019 | None | 0 | None | 0 |
| compas, gender | Fairness | entropy | random | 19 | 5 | 5 | 0.01886393677 | sqrt | 2019 | None | 0 | None | 0.763010874 |
| compas, race | Fairness | entropy | random | None | 5 | 4 | 0.343395585 | sqrt | 2019 | None | 0 | None | 0.390954327 |
| census, gender | Score | gini | best | None | 2 | 1 | 0 | None | 2019 | None | 0 | None | 0 |
| census, race | Score | gini | best | 7 | 4 | 5 | 0 | None | 2019 | None | 0 | None | 0 |
| credit, gender | Score | gini | random | 9 | 2 | 4 | 0 | log2 | 2019 | None | 0 | None | 0 |
| bank, age | Score | gini | best | None | 2 | 1 | 0.446090821 | None | 2019 | None | 0 | None | 0 |
| compas, gender | Score | entropy | best | 9 | 2 | 2 | 0.0399448551 | None | 2019 | None | 0 | None | 0.5589434946 |
| compas, race | Score | entropy | random | 5 | 5 | 5 | 0.06714672935 | None | 2019 | None | 0 | None | 0.645003704 |
| census, gender | Worst | gini | best | None | 2 | 1 | 0 | None | 2019 | None | 0 | None | 0 |
| census, race | Worst | gini | best | 18 | 3 | 3 | 0 | None | 2019 | None | 0 | None | 0 |
| credit, gender | Worst | gini | best | 7 | 2 | 3 | 0 | sqrt | 2019 | None | 0 | None | 0 |
| bank, age | Worst | gini | best | None | 5 | 1 | 0 | None | 2019 | None | 0 | None | 0 |
| compas, gender | Worst | gini | best | 11 | 5 | 2 | 0.403645002 | auto | 2019 | None | 0 | None | 0.004077913416 |
| compas, race | Worst | gini | best | 13 | 4 | 5 | 0.1508125665 | sqrt | 2019 | None | 0 | None | 0.1357372627 |

Table B.2: Logistic Regression Model Hyperparameters

| dataset | optimal | penalty | dual | tol | C | fit_intercept | intercept_scaling | solver | max_iter | multi_class | l1_ratio | random_state | class_weight | verbose | warm_start | n_jobs |
|---|---|---|---|---|---|---|---|---|---|---|---|---|---|---|---|---|
| census, gender | Fairness | l1 | FALSE | 0.3063690768 | 45.23735331 | TRUE | 3.912459476 | saga | 1010 | auto | None | 2019 | None | 0 | FALSE | None |
| census, race | Fairness | l2 | FALSE | 0.5455322919 | 4.993769241 | FALSE | 0.3341263337 | liblinear | 1015 | ovr | None | 2019 | None | 0 | TRUE | None |
| credit, gender | Fairness | l2 | FALSE | 0.4740247231 | 1.397173381 | TRUE | 5.600725761 | sag | 989 | auto | None | 2019 | None | 0 | FALSE | None |
| bank, age | Fairness | l1 | FALSE | 0.2571872535 | 41.35040355 | FALSE | 7.402887921 | liblinear | 1021 | ovr | None | 2019 | None | 0 | FALSE | None |
| compas, gender | Fairness | l2 | FALSE | 0.0001 | 10 | TRUE | 1 | lbfgs | 1000 | ovr | None | 2019 | None | 0 | FALSE | None |
| compas, race | Fairness | l2 | FALSE | 0.7686757028 | 99.9510780 | TRUE | 3.215575399 | newton-cg | 967 | multinomial | None | 2019 | None | 0 | FALSE | None |
| census, gender | Score | l1 | FALSE | 0.1441222156 | 22.30673591 | TRUE | 6.945152409 | liblinear | 988 | auto | None | 2019 | None | 0 | TRUE | None |
| census, race | Score | l2 | FALSE | 0.3265133692 | 61.84453228 | TRUE | 3.874932401 | sag | 1033 | auto | None | 2019 | None | 0 | TRUE | None |
| credit, gender | Score | l2 | FALSE | 0.5381666871 | 54.85753506 | TRUE | 8.0468201 | newton-cg | 985 | multinomial | None | 2019 | None | 0 | FALSE | None |
| bank, age | Score | l2 | FALSE | 0.0001 | 10 | FALSE | 1 | lbfgs | 1000 | ovr | None | 2019 | None | 0 | FALSE | None |
| compas, gender | Score | l2 | FALSE | 0.0001 | 10 | TRUE | 1 | lbfgs | 1000 | ovr | None | 2019 | None | 0 | FALSE | None |
| compas, race | Score | l2 | FALSE | 0.7686757028 | 99.9510780 | TRUE | 3.215575399 | newton-cg | 967 | multinomial | None | 2019 | None | 0 | FALSE | None |
| census, race | Worst | l2 | FALSE | 0.5684828555 | 10.6266278 | FALSE | 8.279780679 | sag | 1002 | multinomial | None | 2019 | None | 0 | TRUE | None |
| census, race | Worst | l1 | FALSE | 0.30588973 | 83.16547417 | FALSE | 6.7588283 | liblinear | 988 | auto | None | 2019 | None | 0 | TRUE | None |
| credit, gender | Worst | l2 | FALSE | 0.5345389535 | 15.71657096 | TRUE | 5.637660656 | lbfgs | 950 | auto | None | 2019 | None | 0 | TRUE | None |
| bank, age | Worst | l2 | FALSE | 0.8079471204 | 59.75685297 | FALSE | 7.781969559 | sag | 956 | multinomial | None | 2019 | None | 0 | TRUE | None |
| compas, gender | Worst | l2 | FALSE | 0.7286915762 | 60.9074889 | FALSE | 4.151139651 | liblinear | 1023 | ovr | None | 2019 | None | 0 | FALSE | None |
| compas, race | Worst | l1 | FALSE | 0.5954089764 | 49.86714429 | FALSE | 6.950411052 | liblinear | 952 | ovr | None | 2019 | None | 0 | FALSE | None |

Table B.3: Random Forest Model Hyperparameters

| dataset | optimal | n_estimators | criterion | max_depth | min_samples_split | min_samples_leaf | min_weight_fraction_leaf | max_features | max_leaf_nodes | min_impurity_decrease | bootstrap | oob_score | warm_start | ccp_alpha | max_samples | random_state | verbose | n_jobs |
|---|---|---|---|---|---|---|---|---|---|---|---|---|---|---|---|---|---|---|
| census, gender | Worst | 90 | log_loss | None | 4 | 2 | 0.0714384564 | log2 | None | 0 | TRUE | TRUE | FALSE | 57 | None | None | 0 | None |
| census, gender | Score | 89 | log_loss | None | 5 | 2 | 0.0005484677352 | auto | None | 0 | TRUE | FALSE | TRUE | 93 | None | 2019 | 0 | None |
| census, gender | Fairness | 57 | log_loss | None | 7 | 4 | 0.096207585549 | sqrt | None | 0 | TRUE | FALSE | TRUE |  | None | None | 0 | None |
| census, race | Worst | 59 | log_loss | None | 4 | 2 | 0.063034064527 | sqrt | None | 0 | TRUE | TRUE | FALSE |  | None | 2019 | 0 | None |
| census, race | Score | 99 | gini | 20 | 7 | 2 | 0.0005404346919 | sqrt | None | 0 | TRUE | FALSE | TRUE |  | None | 2019 | 0 | None |
| census, race | Fairness | 66 | entropy | 15 | 5 | 3 | 0.0064518394 | None | None | 0 | TRUE | FALSE | FALSE |  | None | 2019 | 0 | None |
| credit, gender | Worst | 55 | gini | 11 | 3 | 1 | 0.0275560916 | auto | None | 0 | TRUE | TRUE | FALSE | 89 | None | None | 0 | None |
| credit, gender | Score | 58 | gini | 10 | 3 | 4 | 7.70E-05 | sqrt | None | 0 | TRUE | FALSE | FALSE |  | None | 2019 | 0 | None |
| credit, gender | Fairness | 93 | log_loss | None | 8 | 3 | 0.096044978251 | sqrt | None | 0 | TRUE | TRUE | TRUE |  | None | 2019 | 0 | None |
| bank, age | Worst | 73 | entropy | None | 2 | 1 | 0.00656923715 | None | None | 0 | TRUE | TRUE | FALSE |  | None | 2019 | 0 | None |
| bank, age | Score | 100 | gini | 11 | 2 | 1 | 0 | None | None | 0 | TRUE | TRUE | FALSE |  | None | 2019 | 0 | None |
| bank, age | Fairness | 50 | gini | 18 | 3 | 4 | 0.0472732655 | log2 | None | 0 | TRUE | TRUE | FALSE | 90 | None | None | 0 | None |
| compas, gender | Worst | 100 | gini | None | 2 | 1 | 0 | None | None | 0 | TRUE | FALSE | TRUE |  | None | 2019 | 0 | None |
| compas, gender | Score | 100 | gini | None | 2 | 1 | 0.027623541 | None | None | 0 | TRUE | FALSE | TRUE | 59 | None | 2019 | 0 | None |
| compas, gender | Fairness | 100 | gini | None | 2 | 1 | 0.027623541 | None | None | 0 | TRUE | TRUE | TRUE |  | None | 2019 | 0 | None |
| compas, race | Worst | 100 | gini | None | 2 | 1 | 0 | sqrt | None | 0 | TRUE | TRUE | TRUE |  | None | 2019 | 0 | None |
| compas, race | Score | 57 | gini | None | 9 | 4 | 0.037395834 | sqrt | None | 0 | TRUE | FALSE | FALSE |  | None | 2019 | 0 | None |
| compas, race | Fairness | 57 | gini | None | 9 | 4 | 0.037395834 | sqrt | None | 0 | TRUE | FALSE | FALSE |  | None | 2019 | 0 | None |

Table B.4: Suport Vector Machine Model Hyperparameters

| dataset | optimal | C | kernel | degree | gamma | coef0 | shrinking | probability | tol | cache_size | class_weight | verbose | max_iter | decision_function_shape | break_ties | random_state |
|---|---|---|---|---|---|---|---|---|---|---|---|---|---|---|---|---|
| census_gender | Fairness | 1.254855099 | linear | 7 | scale | 1.688182092 | FALSE | TRUE | 0.2486331443 | 234.4690896 | balanced | 0 | 971 | ovr | TRUE | 2019 |
| census_race | Fairness | 0.0005944318107 | linear | 2 | auto | 4.362727653 | TRUE | TRUE | 0.2784471138 | 248.4022564 | balanced | 0 | 969 | ovo | FALSE | 2019 |
| credit_gender | Fairness | 1.628185921 | linear | 3 | scale | 3.562504195 | FALSE | TRUE | 0.9642535597 | 225.7841529 | balanced | 0 | 1028 | ovr | FALSE | 2019 |
| bank_age | Fairness | 1.341990803 | linear | 0 | auto | 8.379271494 | TRUE | TRUE | 0.4655636744 | 200.5390608 | balanced | 0 | 1005 | ovo | FALSE | 2019 |
| compas_gender | Fairness | 1 | linear | 3 | scale | 0 | TRUE | TRUE | 0.001 | 250 | None | 0 | 1000 | ovr | FALSE | 2019 |
| compas_race | Fairness | 1 | linear | 3 | scale | 0 | TRUE | TRUE | 0.001 | 250 | None | 0 | 1000 | ovr | FALSE | 2019 |
| census_gender | Score | 1.254855099 | linear | 7 | scale | 1.688182092 | FALSE | TRUE | 0.2486331443 | 234.4690896 | balanced | 0 | 971 | ovr | TRUE | 2019 |
| census_race | Score | 2.455335736 | linear | 3 | auto | 5.882134131 | TRUE | TRUE | 0.2567407506 | 216.3447828 | balanced | 0 | 977 | ovo | FALSE | 2019 |
| credit_gender | Score | 1 | linear | 0 | scale | 0 | TRUE | TRUE | 0.001 | 250 | None | 0 | 1000 | ovr | FALSE | 2019 |
| bank_age | Score | 3.55483917 | linear | 9 | auto | 2.135759165 | FALSE | TRUE | 0.2023762998 | 254.3261852 | None | 0 | 965 | ovr | TRUE | 2019 |
| compas_gender | Score | 1 | linear | 3 | scale | 0 | TRUE | TRUE | 0.001 | 250 | None | 0 | 1000 | ovr | FALSE | 2019 |
| compas_race | Score | 1 | linear | 3 | scale | 0 | TRUE | TRUE | 0.001 | 250 | None | 0 | 1000 | ovr | FALSE | 2019 |
| census_gender | Worst | 4.095055553 | linear | 10 | scale | 5.912251425 | FALSE | TRUE | 0.5409397753 | 229.2741173 | balanced | 0 | 1022 | ovr | FALSE | 2019 |
| census_race | Worst | 2.802048724 | linear | 6 | scale | 5.694737901 | FALSE | TRUE | 0.7818051683 | 268.6986856 | balanced | 0 | 955 | ovr | FALSE | 2019 |
| credit_gender | Worst | 1 | linear | 3 | scale | 0 | TRUE | TRUE | 0.001 | 250 | None | 0 | 1000 | ovr | FALSE | 2019 |
| bank_age | Worst | 3.389007545 | linear | 1 | auto | 1.818585713 | FALSE | TRUE | 0.5485891593 | 217.9204432 | balanced | 0 | 971 | ovr | FALSE | 2019 |
| compas_gender | Worst | 1 | linear | 3 | scale | 0 | TRUE | TRUE | 0.001 | 250 | None | 0 | 1000 | ovr | FALSE | 2019 |
| compas_race | Worst | 1 | linear | 3 | scale | 0 | TRUE | TRUE | 0.001 | 250 | None | 0 | 1000 | ovr | FALSE | 2019 |

<u>ACADEMIC VITA</u>
NORMEN YU

Email: normenyu@gmail.com

LinkedIn: [www.linkedin.com/in/normen-yu](www.linkedin.com/in/normen-yu) ● Github: [https://github.com/Pennswood/](https://github.com/Pennswood/)

## Education

**The Pennsylvania State University, University Park**          *Aug 2019 - May 2023*

*Schreyer Honors College, Bachelor of Science in Computer Science*
*Minor in Mathematics* (focus on controls, numerical computation, ODE, and PDE)
  - Thesis: "*FairLay-ML: Intuitive Remedies for Unfairness in Data-Driven Social-Critical Algorithms*"
    Thesis supervisor: Dr. Gang Tan; Thesis mentor: Dr. Saeid Tizpaz Niari

---

## Work Experiences

**Sierra Nevada Corporation**

*DevOps Intern*          Aug 2022 - Dec 2022
  - Developed explainable MLOps infrastructure to facilitate rapid and systematic pipelines for future machine-learning-as-a-service government contracts.
  - Utilizing tools including Streamlit, Apache Airflow, Pytorch, Docker Desktop, and AWS

**Collins Aerospace**

*Software Engineering Intern*          *May 2022 - July 2022*
  - Implemented a network filtering algorithm using the Rust programming language.
  - Evaluated Rust on cybersecurity, memory safety, runtime efficiency, and ease of use.
  - Utilized GitLab CI/CD artifacts for Docker unit testing and code coverage report
  - Proposed and led the effort to use Grafana and SQL to visualize filtering analytics.

*Systems Integration Engineering Co-Op*          *June 2021 - Dec 2021*
  - Responsible for the set-up, screen sharing, and lab camera view of a real-time customer demonstration to win the Air Launch Effects (ALE) OTA2 multi-million dollar contract.
  - Integrated UAV autonomous software and manned aircraft human-machine-interface software for autonomous detect, identify, locate, and report mission demonstration.
  - Implemented and documented DIS and protobuf network interfaces across 16 computers.
  - Tested and verified simulation software for software-in-the-loop and hardware-in-the-loop demonstration of the system.
  - Developed interfaces for human operators to directly control simulation (C# widgets) and visualize data from the mission system in real-time (AFSIM).
  - Investigated Git and Ansible continuous integration and deployment methods for continuous autonomous simulation and mission system updates testing.

**Student Space Program Laboratory**

*Software engineer, Electronics Lead*          *Feb 2020 - May 2021*
  - Coordinated with Carnegie Mellon University to provide vacuum cryogenic chamber testing support for their Iris rover launching in 2022 to the moon.
  - Spearheaded the NASA's Big Idea Challenge's Command and Data Handling team to co-author a proposal for a laser-induced breakdown spectroscopy computer system.
  - Awarded $145,933 with the proposal, Science Vision Award during the final presentation
  - Inspired and managed a group of 5 students to prototype and build a computer system to regulate the temperature of the module, receive command and send data to a rover, and synchronize laser firing and spectrometer data recording.
  - [https://www.nasa.gov/feature/langley/artemis-student-challenge-nasa-selects-university-teams-to-build-technologies-for-the-moon-s/](https://www.nasa.gov/feature/langley/artemis-student-challenge-nasa-selects-university-teams-to-build-technologies-for-the-moon-s/)

---

## Projects and Awards

  - NASA Machine Learning Competition "Run-way Functions: Predict Reconfigurations at US Airports" (**4th place, awarded $4K**). Trained a Logistic Regression model with careful consideration for hyperparameter tuning, temporal decay, and custom dimensionality-extension pre-processing on Google Cloud Platform's high performance machines.
  - Penn State 24 hour HackPSU Hackathon, Spring 2022 (**"HackPSU Hackathon Best Hack" and "Best Artificial Intelligence Solution to Benefit Humanity"**). Integrated live traffic camera video feeder, computer vision technology, and Grafana displays to simulate and display anomalies.